\documentclass[letterpaper]{article}
\usepackage[preprint]{aaai2027}
\usepackage[hyphens]{url}
\usepackage{graphicx}
\usepackage{natbib}
\usepackage{caption}
\usepackage{amsmath}
\usepackage{amssymb}
\usepackage{booktabs}
\title{Self-Routing: Parameter-Free Expert Routing from Hidden States}
\author{
Jama Hussein Mohamud\textsuperscript{\rm 1,\rm 2},
Drew Wagner\textsuperscript{\rm 1,\rm 3},
Mirco Ravanelli\textsuperscript{\rm 1,\rm 3}
}
\affiliations{
\textsuperscript{\rm 1}Mila -- Quebec AI Institute \qquad
\textsuperscript{\rm 2}Universite de Montreal \qquad
\textsuperscript{\rm 3}Concordia University
}

\begin{document}

\maketitle

\begin{abstract}
Mixture-of-Experts (MoE) layers increase model capacity by activating only a small subset of experts per token, and typically rely on a learned router to map hidden states to expert assignments. In this work, we ask whether a dedicated learned router is strictly necessary for MoE routing. We propose \emph{Self-Routing}, a parameter-free routing mechanism that uses a designated subspace of the token hidden state directly as expert logits, eliminating the router projection entirely while leaving the rest of the MoE layer unchanged. We evaluate Self-Routing on language modeling across different expert counts and model scales, and on ImageNet-1K classification by comparing it against a standard learned router, random-routing baselines, and dense non-MoE baselines. Our results show that Self-Routing remains competitive with the learned-router baseline while removing all dedicated routing parameters, and yields more balanced expert utilization, with about 17\% higher average normalized routing entropy and no explicit load-balancing loss. On ImageNet-1K with DeiT-S/16, Self-Routing also slightly improves over the corresponding learned-router MoE. These findings suggest that effective MoE routing can emerge from the hidden representation itself without requiring a separate learned router module.

\end{abstract}

\section{Introduction}

Mixture-of-Experts (MoE) layers have become a standard mechanism for increasing model capacity without activating all parameters for every token~\citep{shazeer2017outrageously,fedus2022switch,jiang2024mixtralexperts,deepseekai2025deepseekv3}. By replacing a dense feed-forward block with a set of experts and activating only a small subset per token, MoE models can scale parameter count while keeping the per-token compute relatively small. A central component of this design is the \emph{router}, which maps token representations to expert assignments.

In most modern MoE architectures, routing is performed by a learned projection from the hidden state to expert logits. This design is simple and effective, and has become the default choice in practice. At the same time, it introduces a dedicated routing module in every MoE layer, together with additional parameters, implementation complexity, and often router-specific training considerations such as load balancing. 

In this work, we study whether the hidden representation already exposes enough information to route tokens effectively. We therefore propose a simple alternative that we call \emph{Self-Routing}. Instead of learning a router matrix, Self-Routing uses a designated subspace of the token hidden state directly as expert logits. We study both a last-$N$ coordinate choice, which uses the final $N$ hidden-state coordinates to route among $N$ experts, and fixed random coordinate choices. \emph{Self-Routing} eliminates only the router parameters while preserving the standard top-$k$ dispatch and aggregation mechanism.

Prior work has examined the role of routing in sparse models from several directions, including learned versus random routing~\citep{dikkala2023on}, hash-based routing~\citep{roller2021hash}, and frozen or randomly initialized routers variants~\citep{fan2024empiricalunderstandingmoedesign}. We discuss these connections, along with adjacent parameter-free sparse-routing ideas, in Section~\ref{sec:related_work}.

Our goal is not to argue that learned routers are unnecessary in general, nor to claim a universal replacement for standard MoE routing. Rather, we ask a narrower empirical question: in the MoE settings studied here, how much does a dedicated learned router actually matter? To answer this, we compare Self-Routing against a standard learned router, random routing baselines, and dense non-MoE baselines in language-model experiments across GPT-2 and LLaMA backbones, different expert counts, and different model sizes, as well as an ImageNet-1K classification comparison with DeiT-S/16. We evaluate not only downstream quality, but also routing behavior through expert utilization and load-balancing behavior.

Our results suggest that a dedicated learned router may be less essential than standard practice assumes. Self-Routing remains competitive with the learned-router baseline on several downstream evaluations while removing all router parameters, across the expert-count and model-size settings we test. Beyond task quality, Self-Routing also yields more balanced expert utilization, achieving higher routing entropy across layers without any explicit load-balancing loss. At the same time, the comparison with random routing shows that expert assignment is not arbitrary. Content-aware routing still matters, but it need not necessarily be mediated by a separate learned projection. Our ImageNet-1K result further suggests that the same idea can remain effective in a standard vision classification setting. Taken together, these findings support the view that MoE routing may depend as much on representational organization as on the explicit form of the router itself.

\section{Method}

\subsection{Background: Mixture-of-Experts Routing}

Mixture-of-Experts (MoE) layers replace a dense feed-forward block with a collection of $N$ expert networks together with a routing mechanism that selects which experts process each token. Given a token hidden state $\mathbf{h} \in \mathbb{R}^{H}$, the router produces a score for each expert, and only the top-$k$ experts are activated. This allows the model to increase parameter count while keeping the per-token compute closer to that of a sparse subset of experts rather than all experts~\citep{shazeer2017outrageously,fedus2022switch,jiang2024mixtralexperts}.

Formally, let $\{E_i\}_{i=1}^N$ denote the experts. A standard MoE layer first computes routing logits
\begin{equation}
    \mathbf{z} = g(\mathbf{h}) \in \mathbb{R}^{N},
\end{equation}
where $g$ is the router. The top-$k$ experts according to $\mathbf{z}$ are selected, and their normalized routing weights are obtained by applying a softmax over the selected experts only. If $\mathcal{T}_k(\mathbf{z})$ denotes the indices of the top-$k$ logits, the MoE output is
\begin{equation}
    \mathrm{MoE}(\mathbf{h}) = \sum_{i \in \mathcal{T}_k(\mathbf{z})} p_i(\mathbf{h}) \, E_i(\mathbf{h}),
\end{equation}
where $p_i(\mathbf{h})$ is the normalized routing weight for expert $i$. In practice, the experts are typically independent feed-forward networks with identical architecture.

In most modern MoE models, the router is implemented as a learned linear projection
\begin{equation}
\label{eq:moe_router}
    \mathbf{z} = \mathbf{h} W_r,
\end{equation}
where $W_r \in \mathbb{R}^{H \times N}$ is learned jointly with the rest of the model. This design is simple and effective, and is the default choice in systems such as Switch Transformer and Mixtral~\citep{fedus2022switch,jiang2024mixtralexperts}. However, it also introduces dedicated routing parameters in every MoE layer. At scale, these router parameters are small relative to the full model, but they are not negligible, and they add an additional mechanism that must be learned and tuned.

A central challenge in sparse MoE training is expert imbalance. Without additional regularization, the router may overuse a small subset of experts while leaving others underutilized. Standard MoE training therefore often adds an auxiliary load-balancing objective~\citep{fedus2022switch}. Let $f_i$ denote the fraction of tokens routed to expert $i$, and let $p_i$ denote the mean routing probability assigned to that expert across a batch. A common balancing loss is
\begin{equation}
    \mathcal{L}_{\text{balance}} = N \sum_{i=1}^{N} f_i p_i.
    \label{eq:moe_balance}
\end{equation}
This term encourages the router to distribute traffic more evenly across experts. In practice, the main training objective is the task loss plus, when used, a weighted load-balancing term.

% The standard learned router can be viewed as a linear readout over the token representation. This raises a simple question: does expert routing truly require a dedicated learned projection, or can the hidden state itself already contain enough information to express expert preference? Our method is built around this question. Rather than learning a separate router matrix, we ask whether a model can directly expose routing logits through its own hidden representation.
The standard learned router computes expert logits via the learned linear projection in Equation~\ref{eq:moe_router}. Recent work has shown this projection need not be learned. The model can effectively route through a fixed random projection by adapting the learned hidden states~\citep{fan2024empiricalunderstandingmoedesign}. This raises a structural question of whether the form of the fixed projection matters. We test the simplest case by reading routing logits from chosen $N$-coordinate subsets of the hidden state at each layer.

\subsection{Self-Routing}
We propose \textbf{Self-Routing}, a parameter-free routing mechanism for MoE layers. Instead of computing routing logits using a learned projection $\mathbf{h} W_r$, Self-Routing uses a designated subspace of the hidden state directly. For an MoE layer with $N$ experts, let $S_\ell \subset \{1,\ldots,H\}$ be a fixed set of $N$ routing-coordinate indices for layer $\ell$. Self-Routing reads those coordinates as the expert logits:
\begin{equation}
    \mathbf{z}_{\text{self}}^{(\ell)} =
    \mathbf{h}^{(\ell)}_{S_\ell} \in \mathbb{R}^{N}.
\end{equation}
We consider two choices for $S_\ell$. The first uses the last $N$ coordinates in every MoE layer, giving a simple coordinate-aligned readout across depth. The second is a random-coordinate variant, where each $S_\ell$ is sampled randomly at initialization and then kept fixed. This hidden-state view exposes a broader design space of coordinate-based routers, including different fixed coordinate subsets across layers. The router then applies the same top-$k$ selection and softmax normalization as in a standard MoE layer. The only difference is how the logits are obtained.

Self-Routing removes the explicit router projection, but it does not make routing fixed or handcrafted. Instead, the model is trained end-to-end so that part of the hidden representation can implicitly encode expert preference. In this sense, the method is parameter-free at the router level while still remaining adaptive through the backbone parameters. The designated routing dimensions act as a readable subspace from which expert preference can be extracted without a separate gating module.

Self-Routing leaves the rest of the MoE layer unchanged. Expert architectures, expert outputs, top-$k$ dispatch, and weighted aggregation all remain identical to the standard learned-router formulation. This makes Self-Routing a drop-in replacement for the router in a conventional top-$k$ MoE layer. The method does not require changes to expert networks, dispatch kernels, or the sparse aggregation step. Unlike attention-derived routing methods, it does not require materializing, storing, or aggregating attention matrices. As a result, Self-Routing remains naturally compatible with optimized attention implementations, including FlashAttention kernels, since the routing logits are obtained by a simple slice of the hidden state rather than by extracting statistics from the attention computation.

For a model with depth $L$, hidden dimension $H$, and $N$ experts per layer, a standard linear router adds $LHN$ learned parameters, whereas Self-Routing adds none. The practical appeal is therefore not only the removal of these parameters, but also the simplification of the MoE layer itself, while retaining input-dependent routing through the hidden state.

\section{Experiments}

\begin{figure}[t]
\centering
\includegraphics[width=\linewidth]{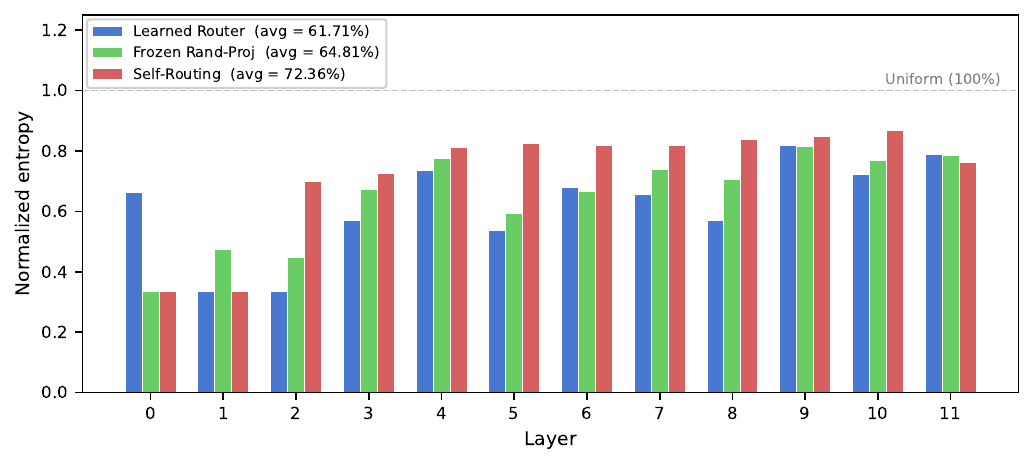}
\caption{\textbf{Per-layer normalized expert-utilization entropy.} Self-Routing achieves higher normalized entropy than the learned and fixed random projection routers across most MoE layers. The dashed line marks normalized entropy 1, corresponding to perfectly uniform usage of all 8 experts.}
\label{fig:entropy_by_layer}
\end{figure}

We evaluate Self-Routing on language modeling and ImageNet-1K classification. For language modeling, our main GPT-2-scale setting uses a 12-layer model with hidden size 768, 12 attention heads, and sequence length 1024, trained on OpenWebText~\citep{Gokaslan2019OpenWeb}. We also evaluate SmolLM-based LLaMA backbones~\citep{allal2024SmolLM} trained on FineWeb-Edu~\citep{penedo2024finewebdatasetsdecantingweb}, using 135M and 360M backbones to test different expert counts and model sizes. For MoE variants, we replace the feed-forward block in every transformer layer with a MoE layer using top-$2$ routing. The GPT-2 MoE variants use 8 experts, while the LLaMA 135M backbone is evaluated with 4 and 8 experts and the LLaMA 360M backbone with 8 experts. All language model variants share the same backbone, expert architecture, optimization recipe, and training budget within each comparison; only the routing mechanism changes. Unless otherwise stated, auxiliary load-balancing loss is disabled in the experiments reported here. 

We compare five pretrained GPT-2 language-model variants. A dense baseline without MoE, a standard learned router, our Self-Routing gate, a fixed random projection, and a random-routing lower bound. We also include the official pretrained GPT-2 checkpoint~\citep{radford2019language} as a reference row. For the LLaMA backbone experiments, we compare learned routing and Self-Routing under matched MoE configurations. For the learned router, routing logits are produced by a learned linear projection. For Self-Routing, we use either the last $N$ hidden-state coordinates or $N$ coordinates sampled independently for each layer and fixed at initialization as routing logits for $N$ experts. For the fixed-random-projection baseline, routing logits are produced by a frozen randomly initialized projection. For the random-routing baseline, fresh random logits are sampled at each forward pass. We evaluate language models on five standard benchmarks using the lm-evaluation-harness~\citep{eval-harness}.

For image classification, we evaluate on ImageNet-1K~\citep{5206848} with a DeiT-S/16 backbone~\citep{touvron2021training}. We first reproduce the standard dense DeiT-S baseline following the recipe of~\citet{touvron2021training}, and then construct DeiT-S MoE variants following the ``Every 2'' strategy used in prior vision MoE work~\citep{videau2025mixtureexpertsimageclassification}, where MoE layers are inserted in every other transformer block of a 12-layer DeiT-S model. The MoE variants use 4 experts with top-$2$ routing. Under this shared recipe, we compare the learned-router MoE model against the corresponding Self-Routing variant, and report the dense baseline together with both MoE models.

\section{Results}

\begin{figure*}[t]
\centering
\includegraphics[width=\linewidth]{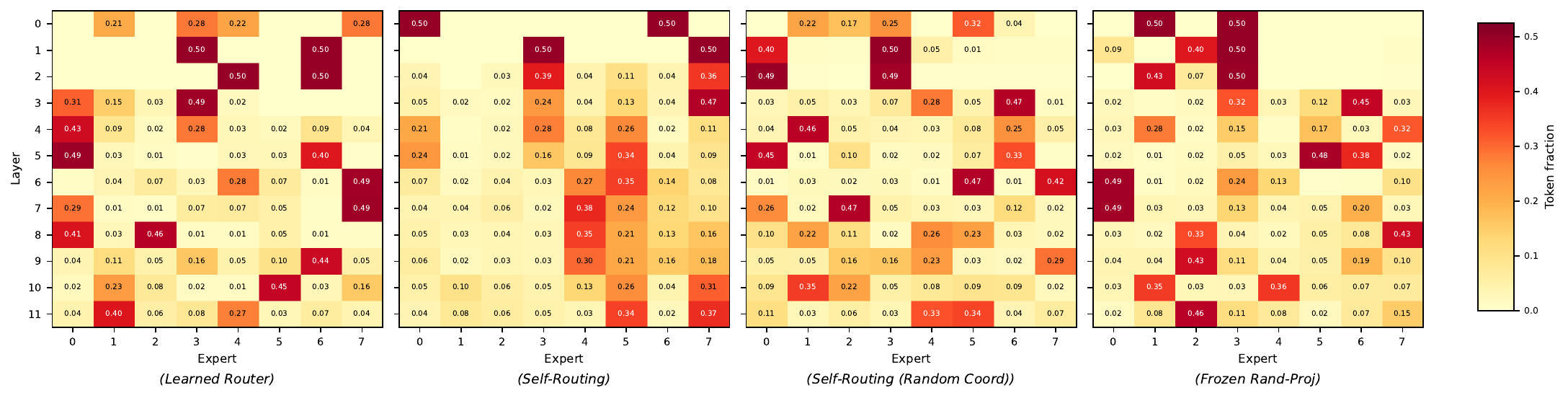}
\caption{\textbf{Layer-by-expert routing fractions.} Each row is a layer and each column is an expert. Hotter colors indicate a larger fraction of routed tokens. Both Self-Routing variants exhibit more even allocation patterns after the earliest layers, while the learned router and fixed random projection show stronger expert concentration and more inactive experts. The last-$N$ coordinate choice shows visible cross-layer continuity in expert usage, whereas random coordinates disrupt this continuity because the routing coordinates differ across layers.}
\label{fig:expert_heatmap}
\end{figure*}

\subsection{Language modeling}
\begin{table}[t]
\centering
\caption{\textbf{Language-model evaluation on GPT-2 Small trained on OpenWebText.} All MoE variants use the same backbone, 8 experts per layer, and top-$2$ routing ($\approx$521M MoE params). Rtr. Params denotes router parameters. ``HF GPT-2'' denotes the official pretrained checkpoint~\citep{radford2019language} from Hugging Face. Accuracy (\%) $\uparrow$ for HellaSwag (HS)~\citep{HellaSwag}, PIQA~\citep{PIQA}, and WinoGrande~\citep{WinoGrande}. Perplexity $\downarrow$ for LAMBADA (LMB)~\citep{LAMBADA} and WikiText (WT).}
\label{tab:moe_language}
\setlength{\tabcolsep}{2pt}
\scriptsize
\begin{tabular}{l c c c c c c}
\toprule
\textbf{Model} & \textbf{Rtr. Params} & \textbf{HS} $\uparrow$ & \textbf{LMB} $\downarrow$ & \textbf{PIQA} $\uparrow$ & \textbf{WT} $\downarrow$ & \textbf{Wino} $\uparrow$ \\
\midrule
HF GPT-2 & --- & 31.1 & 40.1 & 62.9 & 37.4 & 51.6 \\
Dense & --- & 33.3 & 31.6 & 63.0 & 33.6 & 51.7 \\
\midrule
Learned Router & $12 \times H \times N$ & 38.5 & 19.3 & 65.6 & \textbf{26.7} & 53.8 \\
Random & 0 & 32.8 & 35.0 & 59.6 & 34.6 & 52.5 \\
Fixed Rand. & 0 & 38.9 & 20.1 & 65.9 & 27.0 & \textbf{54.1} \\
Self-Route & 0 & \textbf{39.0} & \textbf{18.9} & \textbf{67.4} & 27.4 & 53.8 \\
\bottomrule
\end{tabular}
\end{table}

Table~\ref{tab:moe_language} shows that Self-Routing remains competitive with the standard learned router in the GPT-2 setting. Self-Routing improves over the learned-router baseline on HellaSwag, LAMBADA, and PIQA, matches it on WinoGrande, and trails it slightly on WikiText. The fixed-random-projection baseline is also competitive, suggesting that useful routing signal is already present in the hidden representation.

Random routing performs substantially worse, indicating that the benefit of Self-Routing does not come merely from replacing the learned router with an arbitrary parameter-free mechanism. Taken together, these comparisons suggest that content-dependent routing still matters, but that it need not be implemented through a separate learned projection. Self-Routing remains the simplest competitive alternative, since it avoids both learned routing parameters and an additional fixed projection.
    
\begin{table}[t]
\centering
\caption{\textbf{Routing-coordinate choice.} The random-coordinate Self-Routing variant uses a randomly selected set of $N$ hidden dimensions fixed at initialization. Self-Routing remains competitive with learned routing under both coordinate choices, indicating that the method does not depend on using the last $N$ coordinates specifically.}
\label{tab:coordinate_choice}
\setlength{\tabcolsep}{2.5pt}
\scriptsize
\begin{tabular}{l c c c c c}
\toprule
\textbf{Method} & \textbf{HS} $\uparrow$ & \textbf{LMB} $\downarrow$ & \textbf{PIQA} $\uparrow$ & \textbf{WT} $\downarrow$ & \textbf{Wino} $\uparrow$ \\
\midrule
Learned Router & 38.50 & 19.30 & 65.60 & 26.70 & 53.80 \\
Self-Route (Last $N$) & 39.00 & 18.90 & \textbf{67.40} & 27.40 & 53.80 \\
Self-Route (Rand. $N$) & \textbf{39.61} & \textbf{17.89} & 66.10 & \textbf{26.51} & \textbf{54.14} \\
\bottomrule
\end{tabular}
\end{table}

Table~\ref{tab:coordinate_choice} shows that Self-Routing does not depend on choosing the last $N$ coordinates specifically. The random-coordinate variant remains competitive with both the learned router and the last-$N$ Self-Routing variant across the language-model evaluations.

\begin{figure*}[t]
\centering
\begin{tabular}{@{}cc@{}}
{\scriptsize Layer 4} & {\scriptsize Layer 8} \\[-0.2em]
\includegraphics[width=0.47\linewidth]{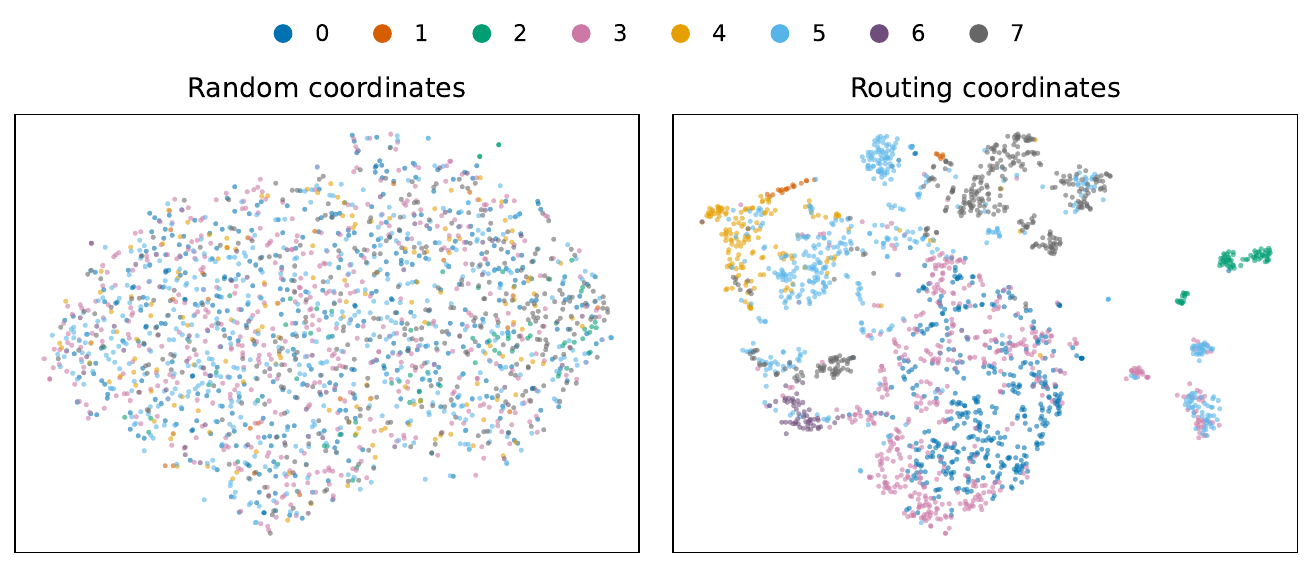} &
\includegraphics[width=0.47\linewidth]{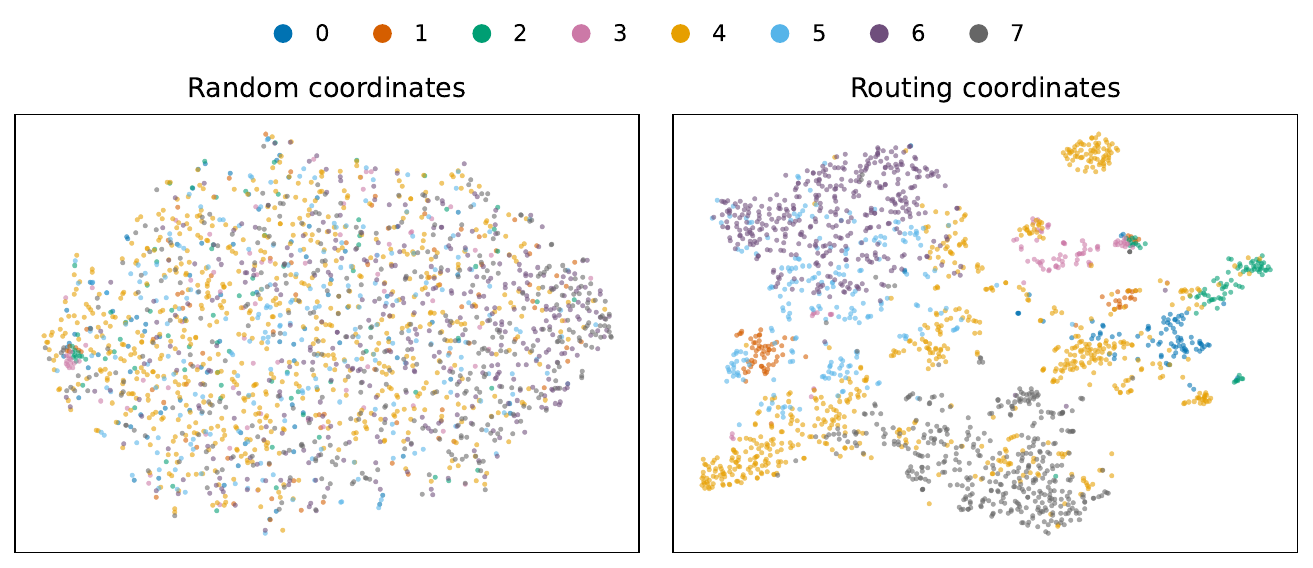} \\[-0.55em]
\multicolumn{2}{c}{\scriptsize Last-$N$ Self-Routing} \\[-0.1em]
\includegraphics[width=0.47\linewidth]{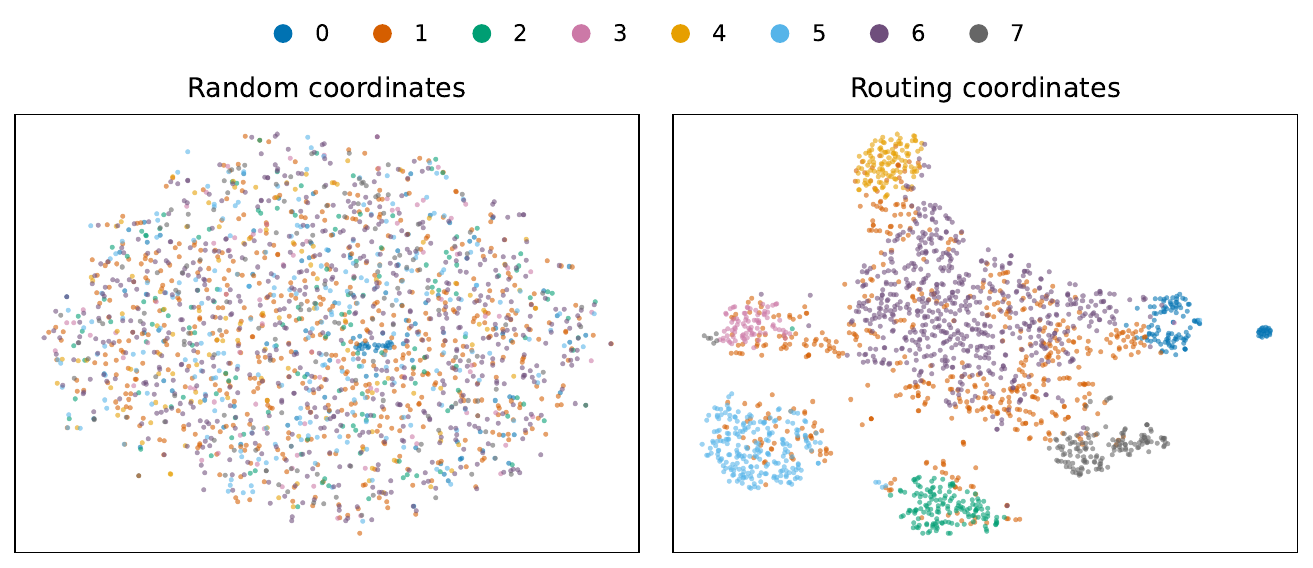} &
\includegraphics[width=0.47\linewidth]{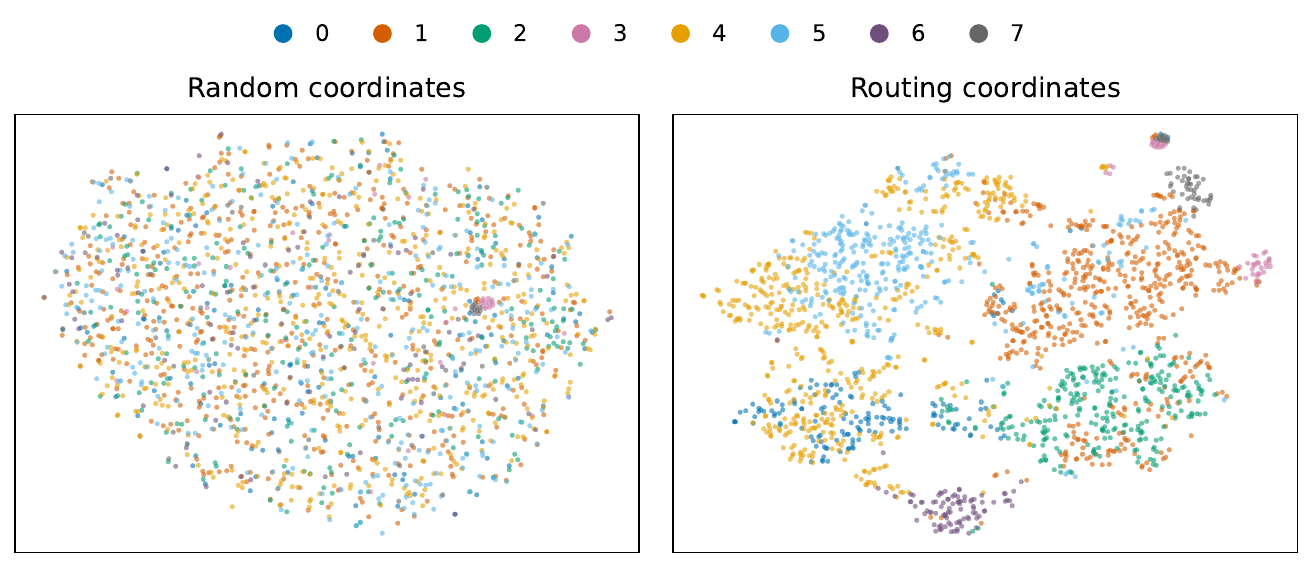} \\[-0.55em]
\multicolumn{2}{c}{\scriptsize Random-coordinate Self-Routing}
\end{tabular}
\caption{\textbf{Routing-coordinate geometry after training.} Layers 4 and 8 of the trained Self-Routing models. In each panel, the left subpanel uses random non-routing coordinates, while the right subpanel uses the coordinates used for routing. Points are colored by the top-1 routed expert. Routing coordinates show clearer expert-separated regions, suggesting that the model places routing-relevant structure into the selected coordinates.}
\label{fig:routing_coordinate_geometry}
\end{figure*}

\begin{table*}[t]
\centering
\caption{\textbf{Expert-count scaling.} Comparison between 4- and 8-expert MoE variants on the LLaMA 135M backbone trained on FineWeb-Edu. MoE Params denotes total parameters after replacing dense MLP blocks with MoE layers; Valid loss is the final validation loss. Bold indicates the better value within each expert-count pair.}
\label{tab:expert_count_scaling}
\vspace{0.2cm}
\scriptsize
\resizebox{\textwidth}{!}{%
\begin{tabular}{llcccccccc}
\toprule
\textbf{Backbone} & \textbf{Router} & \textbf{Experts} & \textbf{MoE Params} & \textbf{Valid loss} $\downarrow$ & \textbf{HellaSwag} $\uparrow$ & \textbf{LAMBADA} $\downarrow$ & \textbf{PIQA} $\uparrow$ & \textbf{Wino.} $\uparrow$ & \textbf{WikiText} $\downarrow$ \\
\midrule
LLaMA 135M & Learned & 4 & 373.46M & \textbf{2.87216} & \textbf{35.21} & \textbf{75.2821} & \textbf{65.34} & 49.09 & \textbf{35.0995} \\
LLaMA 135M & Self-Routing & 4 & 373.39M & 2.87367 & 35.15 & 76.2198 & 63.77 & \textbf{51.22} & 35.1052 \\
\midrule
LLaMA 135M & Learned & 8 & 692.04M & 2.84179 & \textbf{36.16} & 65.0649 & \textbf{65.18} & \textbf{52.33} & \textbf{33.3750} \\
LLaMA 135M & Self-Routing & 8 & 691.90M & \textbf{2.83784} & 36.15 & \textbf{60.7573} & 64.64 & 50.12 & 33.4575 \\
\bottomrule
\end{tabular}
}
\end{table*}

\begin{table*}[t]
\centering
\caption{\textbf{Model-size scaling.} Larger-scale experiments using LLaMA 135M and 360M backbones converted into MoE models and trained on FineWeb-Edu. MoE Params denotes total parameters after replacing dense MLP blocks with MoE layers; Valid loss is the final validation loss. Bold indicates the better value within each backbone pair.}
\label{tab:model_size_scaling}
\vspace{0.2cm}
\scriptsize
\resizebox{\textwidth}{!}{%
\begin{tabular}{llcccccccc}
\toprule
\textbf{Backbone} & \textbf{Router} & \textbf{Experts} & \textbf{MoE Params} & \textbf{Valid loss} $\downarrow$ & \textbf{HellaSwag} $\uparrow$ & \textbf{LAMBADA} $\downarrow$ & \textbf{PIQA} $\uparrow$ & \textbf{Wino.} $\uparrow$ & \textbf{WikiText} $\downarrow$ \\
\midrule
LLaMA 135M & Learned & 8 & 692.04M & 2.84179 & \textbf{36.16} & 65.0649 & \textbf{65.18} & \textbf{52.33} & \textbf{33.3750} \\
LLaMA 135M & Self-Routing & 8 & 691.90M & \textbf{2.83784} & 36.15 & \textbf{60.7573} & 64.64 & 50.12 & 33.4575 \\
\midrule
LLaMA 360M & Learned & 8 & 2.01B & \textbf{2.78152} & \textbf{37.41} & 50.6316 & 64.91 & \textbf{50.12} & \textbf{30.8210} \\
LLaMA 360M & Self-Routing & 8 & 2.01B & 2.78465 & 36.97 & \textbf{50.2209} & \textbf{65.29} & 49.80 & 30.9495 \\
\bottomrule
\end{tabular}
}
\end{table*}

\begin{table*}[t]
\centering
\caption{\textbf{Multi-seed comparisons.} Each row reports mean $\pm$ 95\% CI (std) over three independently trained seeds. Both comparisons use 8 experts: GPT-2 124M ($\approx$521M MoE params) and LLaMA 135M ($\approx$692M MoE params). These multi-seed runs use fewer training tokens than the corresponding main runs; the GPT-2 runs nevertheless remain near the Chinchilla compute-optimal token ratio, roughly 20 training tokens per parameter~\citep{hoffmann2022trainingcomputeoptimallargelanguage}. Accuracy metrics are percentages.}
\label{tab:gpt2_multiseed}
\vspace{0.2cm}
\scriptsize
\resizebox{\textwidth}{!}{%
\begin{tabular}{llcccccc}
\toprule
\textbf{Setting} & \textbf{Router} & \textbf{Valid loss} $\downarrow$ & \textbf{HellaSwag} $\uparrow$ & \textbf{LAMBADA} $\downarrow$ & \textbf{PIQA} $\uparrow$ & \textbf{Wino.} $\uparrow$ & \textbf{WikiText} $\downarrow$ \\
\midrule
GPT-2 124M & Learned & \textbf{2.7109 $\pm$ 0.0104 (0.0042)} & 37.04 $\pm$ 0.64 (0.26) & 21.48 $\pm$ 2.53 (1.02) & 65.14 $\pm$ 2.46 (0.99) & 51.22 $\pm$ 2.41 (0.97) & \textbf{28.09 $\pm$ 1.70 (0.69)} \\
GPT-2 124M & Self-Routing & 2.7118 $\pm$ 0.0123 (0.0050) & \textbf{37.22 $\pm$ 0.78 (0.31)} & \textbf{20.97 $\pm$ 1.39 (0.56)} & \textbf{66.21 $\pm$ 1.63 (0.66)} & \textbf{51.65 $\pm$ 3.04 (1.23)} & 28.60 $\pm$ 1.40 (0.57) \\
\midrule
LLaMA 135M & Fixed random projection & 2.8499 $\pm$ 0.0071 (0.0029) & 35.85 $\pm$ 0.88 (0.35) & 69.06 $\pm$ 7.24 (2.91) & 64.84 $\pm$ 0.74 (0.30) & \textbf{51.28 $\pm$ 5.06 (2.04)} & 34.15 $\pm$ 0.50 (0.20) \\
LLaMA 135M & Self-Routing & \textbf{2.8376 $\pm$ 0.0081 (0.0033)} & \textbf{36.10 $\pm$ 0.85 (0.34)} & \textbf{60.51 $\pm$ 2.72 (1.10)} & \textbf{65.12 $\pm$ 1.33 (0.54)} & 50.72 $\pm$ 1.20 (0.48) & \textbf{33.61 $\pm$ 0.61 (0.25)} \\
\bottomrule
\end{tabular}
}
\end{table*}

\begin{center}
    \includegraphics[width=\linewidth]{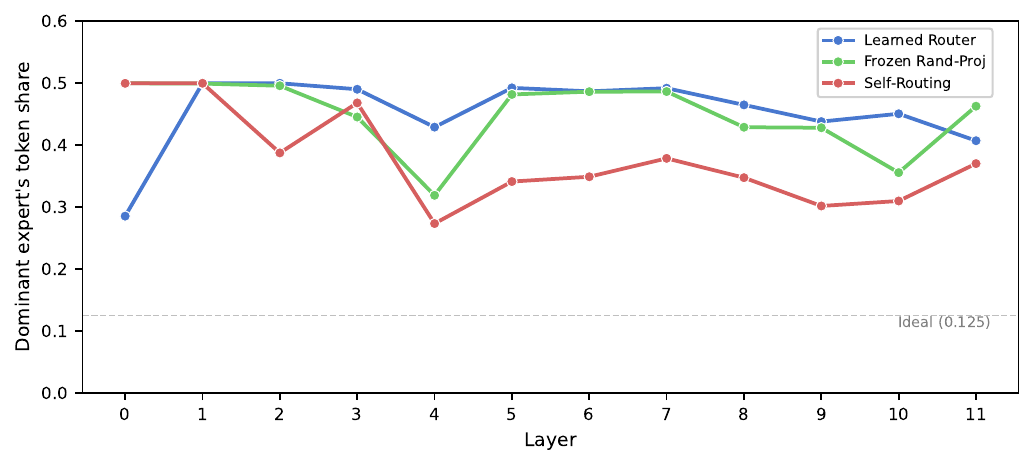}
    \captionof{figure}{\textbf{Maximum expert fraction by layer.} For each layer, we plot the fraction of tokens captured by the single most-used expert. Lower values indicate less concentration. Self-Routing is consistently less dominated by a single expert than either the learned router or fixed random projection.}
    \label{fig:max_expert_fraction}
\end{center}

Tables~\ref{tab:expert_count_scaling} and~\ref{tab:model_size_scaling} extend the language-model comparison to larger LLaMA backbones. Across these runs, Self-Routing stays competitive with learned routing across expert counts and model sizes. Although these experiments use fewer training tokens than the GPT-2 runs, they suggest that the comparison is not specific to a single GPT-2-scale configuration.

To check that these comparisons are not driven by a single run, we also retrain the relevant models across multiple random seeds. Table~\ref{tab:gpt2_multiseed} repeats the GPT-2 comparison with three independently trained seeds. Self-Routing remains similar to the learned router in validation loss, improves the mean on HellaSwag, LAMBADA, PIQA, and WinoGrande, and trails on WikiText. The same table further compares Self-Routing with the fixed random projection baseline across three LLaMA backbone seeds, where Self-Routing improves validation loss and most downstream metrics.

\subsection{Expert utilization}

We examine how evenly each routing mechanism distributes tokens across experts. For this analysis, we collect expert assignments over 4.43M routed tokens and compute, for each MoE layer, the expert fractions $f_i$ and the corresponding Shannon entropy
% \begin{equation}
$\mathcal{H} = - \sum_{i=1}^{N} f_i \log f_i$.
% \end{equation}
We report normalized entropy by dividing by $\log N$, so a value of 1 corresponds to perfectly uniform expert usage.

Figure~\ref{fig:entropy_by_layer} shows that Self-Routing yields consistently higher expert-utilization entropy than both the learned router and the fixed random projection across most layers. Averaged over all 12 MoE layers, normalized entropy increases from 0.617 for the learned router to 0.648 for fixed random projection and 0.724 for Self-Routing. Relative to the learned router, this is a gain of about 17\% for Self-Routing. The strongest imbalance appears in the earliest layers for all three methods. The first two layers are close to the two-expert regime, with normalized entropy near $\log 2 / \log 8 \approx 0.33$. A plausible interpretation is that early hidden representations have not yet differentiated enough to support rich routing decisions, regardless of the routing mechanism. After this initial phase, however, Self-Routing settles into a much more stable high-entropy regime, while the learned router and fixed random projection remain more concentrated and more variable across depth.

The same pattern is visible when inspecting the full expert-allocation matrices. Figure~\ref{fig:expert_heatmap} visualizes, for each layer, the fraction of tokens routed to each expert. The learned router exhibits a patchier allocation pattern with several highly dominant experts and some effectively dead experts. Fixed random projection is slightly more balanced than the learned router in some layers, but still shows strong concentration and persistent expert dominance. In contrast, both Self-Routing variants become substantially more uniform from layer 2 onward, with fewer near-zero columns and fewer abrupt shifts in expert dominance. The last-$N$ Self-Routing variant also shows visually continuous expert-use patterns across layers, while the random-coordinate variant breaks this continuity because each layer routes through a different coordinate subset. This indicates that Self-Routing’s competitiveness is not achieved by collapsing routing onto a small subset of experts; if anything, it is accompanied by broader expert participation.

Figure~\ref{fig:max_expert_fraction} summarizes routing concentration with a single statistic, i.e., the maximum expert fraction in each layer. Lower values indicate that no single expert dominates the routing distribution. Self-Routing stays below both learned routing and fixed random projection in most layers, again showing a more even distribution of traffic. Averaged across layers, the largest expert receives 45.3\% of routed tokens for the learned router, 44.9\% for fixed random projection, and 37.7\% for Self-Routing. Taken together, Figures~\ref{fig:entropy_by_layer}, \ref{fig:expert_heatmap}, and~\ref{fig:max_expert_fraction} suggest that Self-Routing not only matches the learned router on downstream evaluation, but also produces more balanced expert utilization than either learned routing or fixed random projection, without any explicit load-balancing loss.

\subsection{Routing coordinate geometry}

We analyze whether routing structure emerges in the chosen coordinates themselves. Figure~\ref{fig:routing_coordinate_geometry} compares random non-routing coordinates with the routing coordinates in the same trained checkpoint. The routing coordinate view shows clearer expert-separated regions, suggesting that the model learns to place routing-relevant specialization into the designated coordinates rather than relying only on generic LayerNorm-normalized geometry.

We also analyze top-routed tokens on the OpenWebText validation set. For each layer, every token position is assigned to its top-1 expert, and we inspect the tokens routed most often to each expert. Experts show repeated token-group preferences rather than arbitrary assignment patterns, including punctuation and document-boundary tokens, relation and function words, numeric tokens, local continuation tokens, and late-layer prefix-like subwords.

\begin{center}
\includegraphics[width=\linewidth]{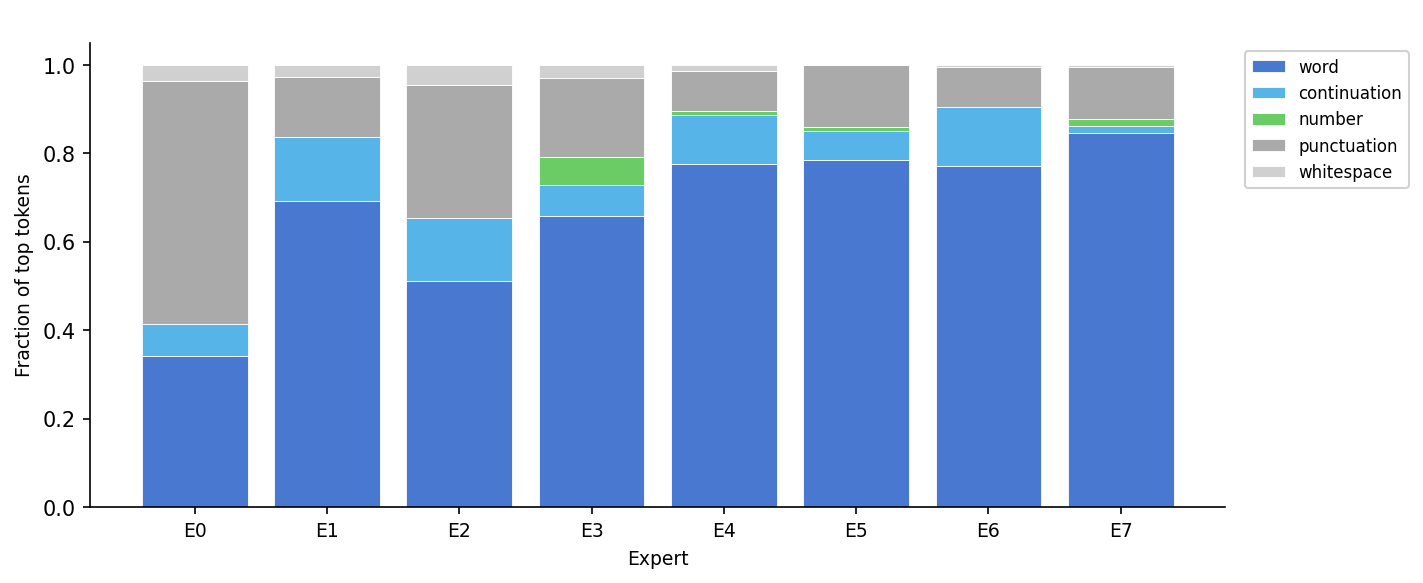}
\captionof{figure}{Token-category mix by expert. Each bar summarizes the top-routed tokens for one expert, averaged over all 12 layers. Experts differ in their mix of word, continuation, number, punctuation, and whitespace tokens.}
\label{fig:token_category_summary}
\end{center}

Figure~\ref{fig:token_category_summary} summarizes these preferences at the category level. The strongest contrast is expert 0, which receives many punctuation and whitespace tokens, while several later experts are dominated by word and continuation tokens. These patterns provide another indication that Self-Routing learns structured expert use from the hidden representation itself. 

The supplementary material provides full layer-wise routing-coordinate visualizations for both last-$N$ and random-coordinate Self-Routing, token-affinity visualizations, and training curves showing similar validation-loss trajectories for Self-Routing and learned routing.

\subsection{Self-Routing with load-balancing}

\begin{table}[t]
\centering
\caption{\textbf{Self-Routing with balancing mechanisms.} LLaMA 135M with 8 experts on FineWeb-Edu. LB denotes the auxiliary load-balancing loss. Bold marks the best value within these Self-Routing variants.}
\label{tab:self_routing_balancing}
\scriptsize
\setlength{\tabcolsep}{2pt}
\begin{tabular}{lcccccc}
\toprule
\textbf{Router} & \textbf{Valid} $\downarrow$ & \textbf{HS} $\uparrow$ & \textbf{LMB} $\downarrow$ & \textbf{PIQA} $\uparrow$ & \textbf{Wino} $\uparrow$ & \textbf{WT} $\downarrow$ \\
\midrule
Self-Routing & 2.83784 & 36.15 & \textbf{60.7573} & \textbf{64.64} & 50.12 & 33.4575 \\
Self-Routing + LB & \textbf{2.83257} & \textbf{36.30} & 62.7297 & 64.53 & \textbf{51.62} & \textbf{33.4203} \\
Self-Routing Expert-Choice & 2.83468 & 36.10 & 64.6120 & 63.77 & 51.46 & 33.5636 \\
\bottomrule
\end{tabular}
\end{table}

Table~\ref{tab:self_routing_balancing} shows that Self-Routing remains compatible with common balancing mechanisms. Adding the auxiliary load-balancing loss from Equation~\ref{eq:moe_balance} gives very similar performance to the base Self-Routing variant, with small improvements on some metrics and small regressions on others. Expert-choice routing takes a different route to balance by letting each expert select its highest-scoring tokens under a capacity constraint, rather than assigning each token only to its top-scoring experts. It also remains close to the base Self-Routing. We therefore view the table mainly as evidence that standard balancing mechanisms do not break Self-Routing's functionality. This is useful because the utilization results above suggest that Self-Routing already tends to spread traffic relatively evenly, while still allowing additional balancing constraints when desired.

\subsection{ImageNet-1K classification}

\begin{table}[t]
\centering
\caption{\textbf{ImageNet-1K comparison on DeiT-S/16.} We reproduce the dense DeiT-S baseline and compare learned-router and Self-Routing MoE variants under the same recipe. MoE variants use 4 experts. Top-1 test accuracy is reported as mean $\pm$ 95\% CI (std.) over three seeds.}
\label{tab:moe_vision_selfroute}
\scriptsize
\setlength{\tabcolsep}{3pt}
\begin{tabular}{l c c c}
\toprule
\textbf{Model} & \textbf{Router} & \textbf{Params} & \textbf{Top-1} $\uparrow$ \\
\midrule
Dense DeiT-S/16 & --- & 22.04M & \textbf{79.80 $\pm$ 0.42 (0.17)} \\
\midrule
DeiT-S/16 MoE & Learned & 43.32M & 79.54 $\pm$ 1.20 (0.48) \\
DeiT-S/16 MoE & Self-Routing & 43.31M & \underline{79.71 $\pm$ 0.45 (0.18)} \\
\bottomrule
\end{tabular}
\end{table}

Table~\ref{tab:moe_vision_selfroute} shows the results on ImageNet-1K across three seeds with 4 experts, using the Every 2 strategy described in \citet{videau2025mixtureexpertsimageclassification}. That study shows that ImageNet MoE does not redefine state-of-the-art ImageNet performance despite the added complexity. Nevertheless, in this setup, Self-Routing remains competitive with the corresponding learned-router MoE, and in this run attains a modestly higher top-1 accuracy. The results suggest that the central idea is not restricted to language modeling and can also remain effective in a standard vision classification setting.

\FloatBarrier
\section{Related work}
\label{sec:related_work}

Mixture-of-Experts has long been studied as a form of conditional computation, where only a subset of model parameters is activated for each input~\citep{shazeer2017outrageously}. In large language models, MoE is primarily used as a scaling mechanism. By routing each token to a small number of experts, models can increase total parameter count without increasing per-token compute proportionally~\citep{fedus2022switch,jiang2024mixtralexperts,deepseekai2025deepseekv3}. In these systems, routing is typically implemented by a learned projection from the hidden state to expert logits, and this design has become the default choice in practice.

Several works have examined how important the router is to MoE performance. \citet{dikkala2023on} argue that learning to route provides a meaningful advantage over data-independent routing, and provide both theoretical and empirical evidence that a trainable router can learn structured partitions of the input space. \citet{fan2024empiricalunderstandingmoedesign} further study MoE design choices empirically, including router type and routing granularity, and report that frozen or randomly initialized routers can remain competitive in some GPT-2-scale settings. \citet{zoph2022stmoedesigningstabletransferable} study how to make large sparse MoE models stable and transferable in practice, highlighting router-stabilization techniques and broader routing and load-balancing design considerations for large-scale MoE training.

Other work has explored alternatives that weaken or remove the dependence on a fully learned router. \citet{roller2021hash} propose hash-based routing as a simple non-learned sparse assignment mechanism, while \citet{chen2023sparsemoenewdropout} show that sparse MoE models with fixed random routing can still be effective in some settings. \citet{pmlr-v139-lewis21a} replace standard token-wise routing with a balanced assignment procedure that avoids auxiliary load-balancing losses, while \citet{zhou2022expertchoice} modify the routing direction itself by letting experts select tokens. \citet{puigcerver2024softmoe} replace hard token-to-expert assignment with a fully differentiable soft mixing scheme. Adjacent sparse-routing work has also considered parameter-free routing in Mixture-of-Depths by deriving token-importance signals from attention statistics rather than a learned router~\citep{gadhikar2025attention}. In contrast, Self-Routing operates in MoE and reads per-expert logits directly from hidden states, without requiring access to attention maps. Our method is closest in spirit to parameter-free routing approaches, but differs in retaining input-dependent routing without a separate learned router module.

Taken together, these works suggest that the role of routing in sparse models is richer than the standard learned-projection formulation might imply. Our contribution is to revisit this question specifically for expert routing in MoE language models, and to study whether a parameter-free but still input-dependent router can remain competitive with the standard learned alternative.

\section{Conclusion}

% Our study revisits a basic assumption in modern MoE design: that expert routing must be mediated by a dedicated learned projection. Across the language-modeling and ImageNet-1K settings studied here, Self-Routing shows that this is not always necessary. By reading routing logits directly from a designated hidden-state subspace, Self-Routing removes all router parameters while remaining competitive with a standard learned router on downstream evaluation. The comparison with random routing further indicates that routing quality still matters: the key point is not that expert assignment can be arbitrary, but that useful input-dependent routing can emerge from the hidden representation itself without a separate routing module. Beyond task metrics, Self-Routing also produces more balanced expert utilization, with higher routing entropy and lower concentration on the most-used expert despite using no explicit load-balancing loss.

Our study revisits a standard assumption in modern MoE design: that expert routing requires learned, parameterized routers. In the settings studied here, Self-Routing shows this is not always necessary. By designating a fixed coordinate-aligned subspace of the hidden state as routing logits, Self-Routing achieves competitive performance on most evaluation tasks with zero router parameters. Beyond task metrics, Self-Routing also produces more balanced expert utilization, with higher routing entropy and lower concentration on the most-used expert despite using no explicit load-balancing loss.

\section{Limitations and Future Work}
\label{sec:limitation_and_future_work}
% Our study focuses on two settings: one main GPT-2-scale language-model configuration and one DeiT-S/16 vision configuration. This scope is sufficient to test the core question of whether a dedicated learned router is necessary in the MoE settings we study, but broader validation across larger scales, different numbers of experts, and more demanding routing regimes remains an important next step. We also evaluate a single routing-subspace choice, namely the last $N$ hidden dimensions. While this simple readout works well here, it will be valuable to test how robust the behavior is across alternative designated subspaces and across other sparse architectures and modalities.

% These limitations point naturally to future work. The most immediate next steps are to broaden the routing baselines, extend the evaluation across scales, and study whether the designated routing subspace exhibits consistent specialization across training. More broadly, we hope this work encourages a shift in perspective: instead of treating the router as an unquestioned architectural primitive, it may be fruitful to ask when routing can be read out from representations that the model already learns for other purposes.

Our study considers GPT-2-scale and LLaMA backbone models with different expert counts and model sizes, and includes ImageNet-1K with DeiT-S/16 as a secondary domain. This scope is sufficient to demonstrate that a standard learned linear router may not always be necessary. What remains to be seen is whether our observations will hold across broader model families, larger scales, and more challenging domains. Further investigation is also needed to understand whether the benefits of Self-Routing come primarily from aligning the routing logits to hidden-state coordinates, from using the same coordinates across all layers, or from simply fixing a coordinate subset at initialization. The fixed random projection baseline (Table~\ref{tab:moe_language}) suggests that using the same routing subspace across layers provides benefits beyond simply avoiding learned parameters, though the mechanisms underlying this difference warrant further investigation. We conjecture that the global and coordinate-aligned routing of Self-Routing may make optimization easier, by allowing a consistent notion of "specialization" across layers and throughout training. We hope to inspire future work in this direction.

\section*{Acknowledgements}

We thank Francesco Bonzi, Mohsin Hasan, Wenhao Huang, and Pascal Tikeng for helpful discussions during the development of this project. The first author is grateful to Yoshua Bengio for supporting and funding his PhD research. This research was enabled in part by computational resources provided by the Digital Research Alliance of Canada\footnote{\url{https://alliancecan.ca}} and Mila\footnote{\url{https://mila.quebec}}.

\bibliography{selfrouting}

\clearpage
\appendix
\section{Supplementary Material}

\subsection{Load-Balancing for the Learned-Router}
\label{app:balancing_mechanisms}

We additionally evaluate two standard balancing mechanisms for learned routing in the 8-expert LLaMA 135M setting. The first adds the auxiliary load-balancing objective from Equation~\ref{eq:moe_balance}. The second uses expert-choice routing, where each expert selects its highest-scoring tokens subject to a capacity constraint. These variants keep the same backbone, dataset, expert count, and training budget as the corresponding LLaMA comparison.

\begin{center}
\captionof{table}{Load-balancing and expert-choice for the learned-router baseline using LLaMA 135M backbone with 8 experts on FineWeb-Edu. LB denotes the auxiliary load-balancing loss. Bold marks the best value in each column.}
\label{tab:learned_balancing_mechanisms}
\scriptsize
\setlength{\tabcolsep}{2pt}
\begin{tabular}{lcccccc}
\toprule
\textbf{Router} & \textbf{Valid} $\downarrow$ & \textbf{HS} $\uparrow$ & \textbf{LMB} $\downarrow$ & \textbf{PIQA} $\uparrow$ & \textbf{Wino} $\uparrow$ & \textbf{WT} $\downarrow$ \\
\midrule
Learned & 2.84179 & 36.16 & 65.0649 & 65.18 & \textbf{52.33} & 33.3750 \\
Learned + LB & 2.82117 & 36.31 & \textbf{62.9689} & \textbf{66.16} & 52.01 & 33.0048 \\
Learned Expert-Choice & \textbf{2.81758} & \textbf{36.52} & 64.7940 & 64.96 & 50.67 & \textbf{32.5575} \\
\bottomrule
\end{tabular}
\end{center}

Table~\ref{tab:learned_balancing_mechanisms} shows that both balancing mechanisms behave as expected for the learned-router baseline under this limited training budget, with modest improvements in validation loss and mixed downstream changes.

\subsection{Routing Coordinate Geometry}
\label{app:routing_coordinate_geometry}

\begin{figure*}[t]
\centering
\begin{minipage}{0.395\textwidth}
\centering
\includegraphics[width=\linewidth]{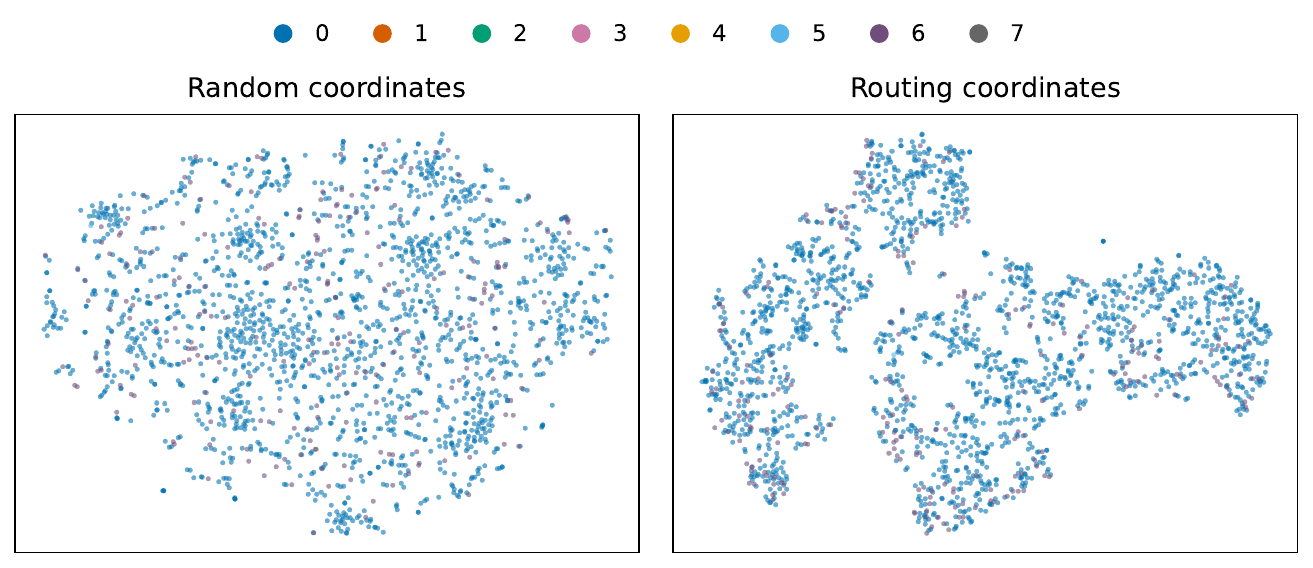}\\[-0.25em]
\scriptsize Layer 0
\end{minipage}
\begin{minipage}{0.395\textwidth}
\centering
\includegraphics[width=\linewidth]{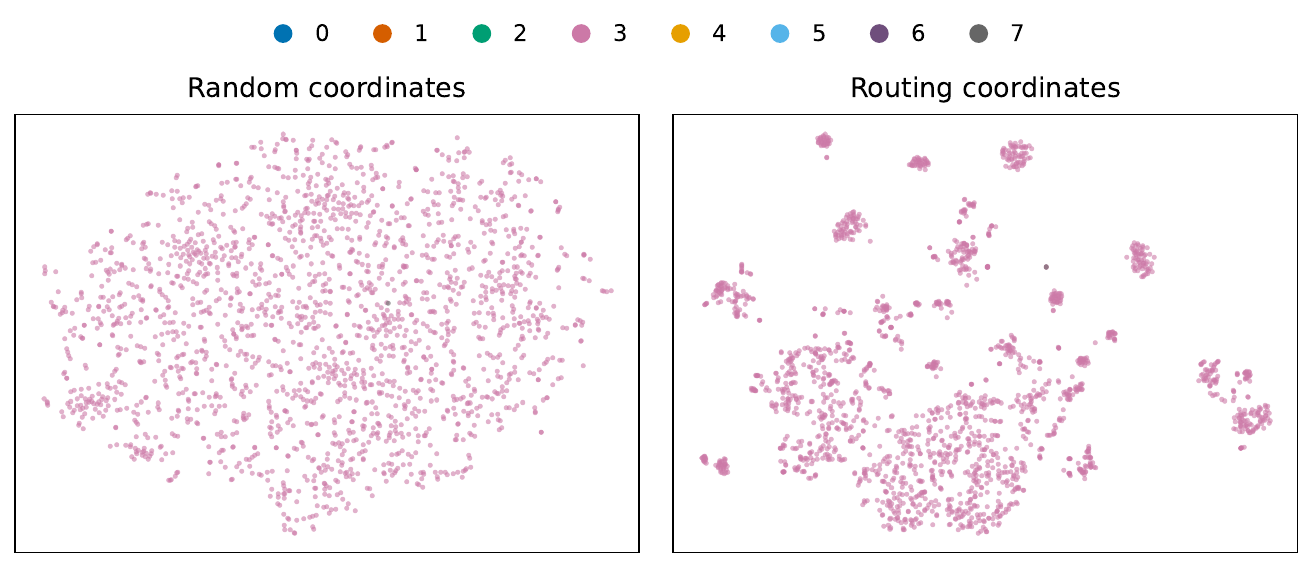}\\[-0.25em]
\scriptsize Layer 1
\end{minipage}

\vspace{0.12em}

\begin{minipage}{0.395\textwidth}
\centering
\includegraphics[width=\linewidth]{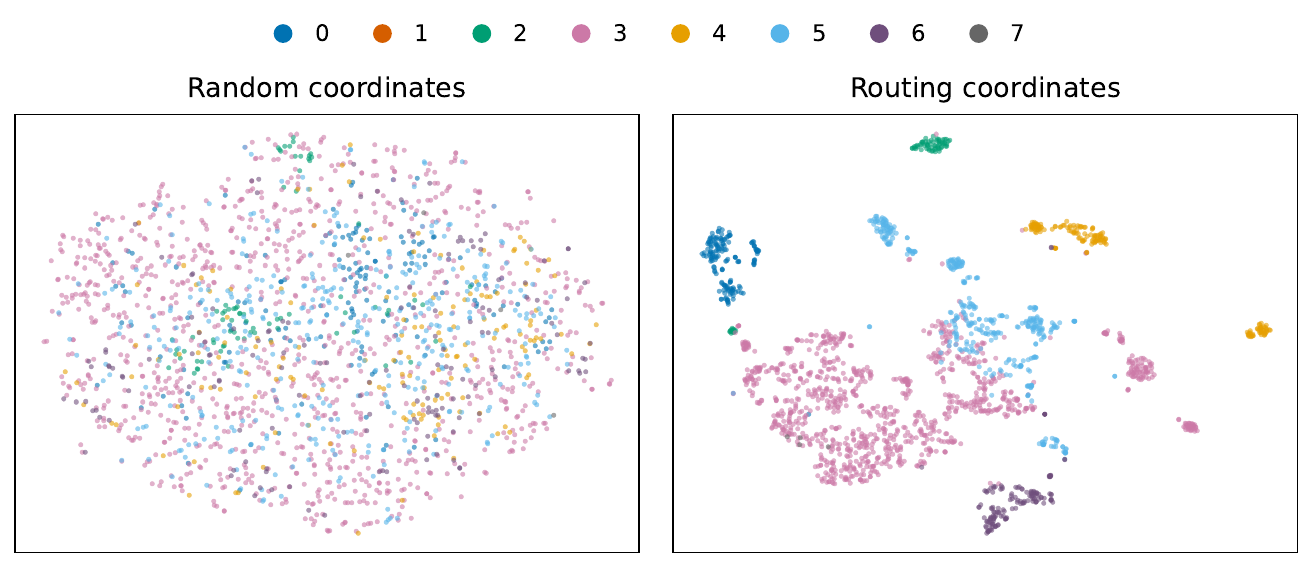}\\[-0.25em]
\scriptsize Layer 2
\end{minipage}
\begin{minipage}{0.395\textwidth}
\centering
\includegraphics[width=\linewidth]{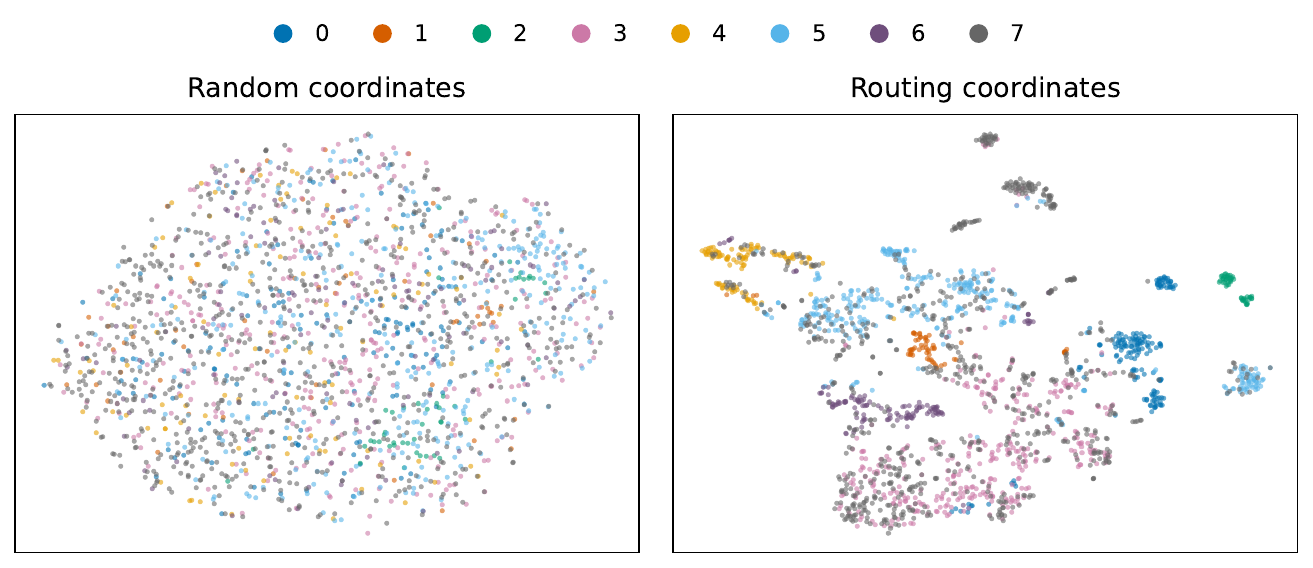}\\[-0.25em]
\scriptsize Layer 3
\end{minipage}

\vspace{0.12em}

\begin{minipage}{0.395\textwidth}
\centering
\includegraphics[width=\linewidth]{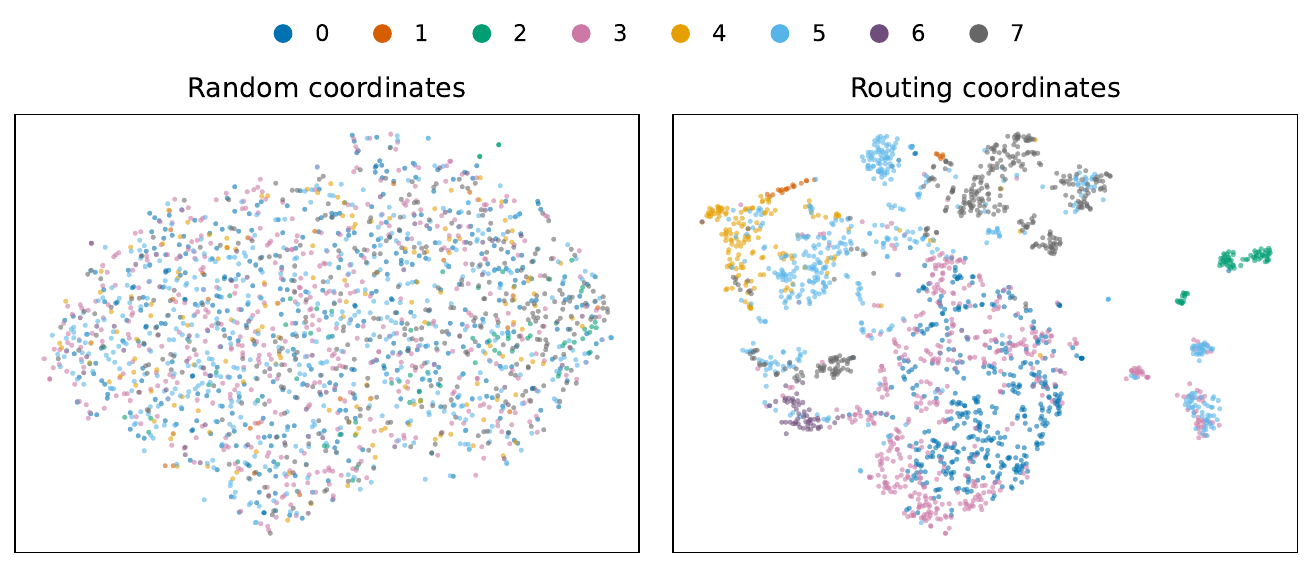}\\[-0.25em]
\scriptsize Layer 4
\end{minipage}
\begin{minipage}{0.395\textwidth}
\centering
\includegraphics[width=\linewidth]{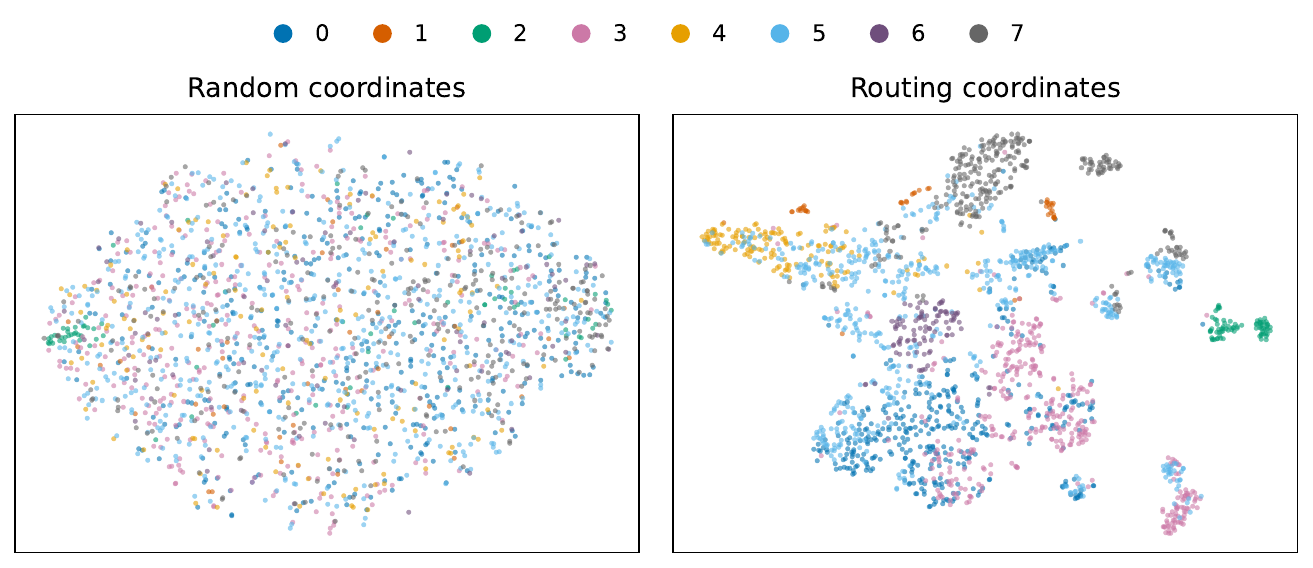}\\[-0.25em]
\scriptsize Layer 5
\end{minipage}

\vspace{0.12em}

\begin{minipage}{0.395\textwidth}
\centering
\includegraphics[width=\linewidth]{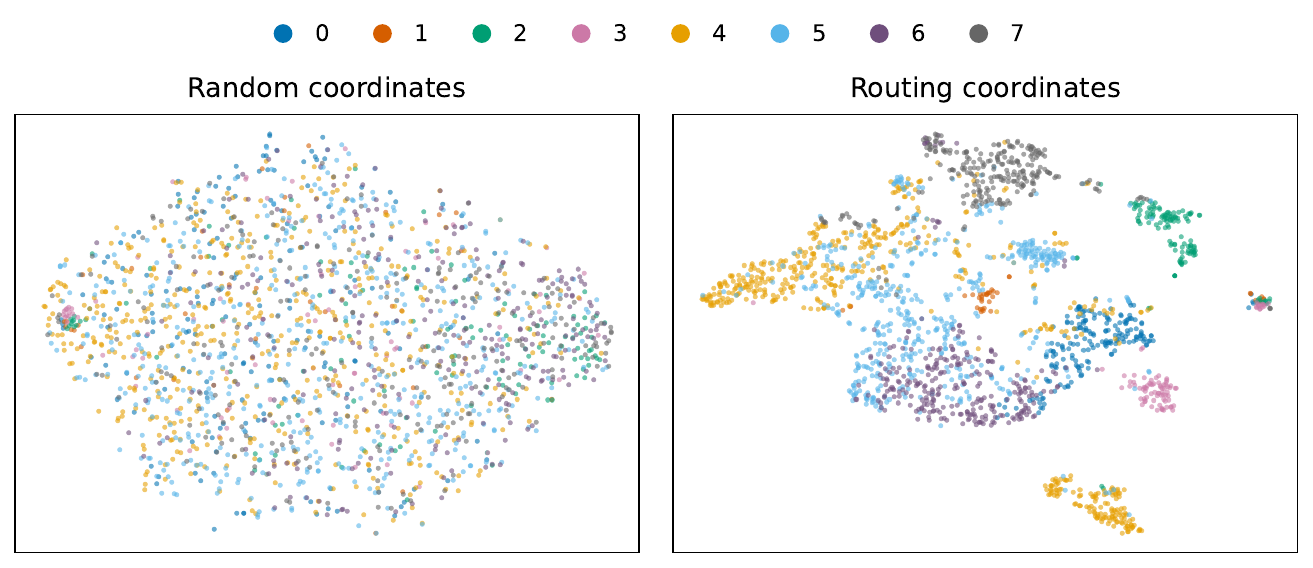}\\[-0.25em]
\scriptsize Layer 6
\end{minipage}
\begin{minipage}{0.395\textwidth}
\centering
\includegraphics[width=\linewidth]{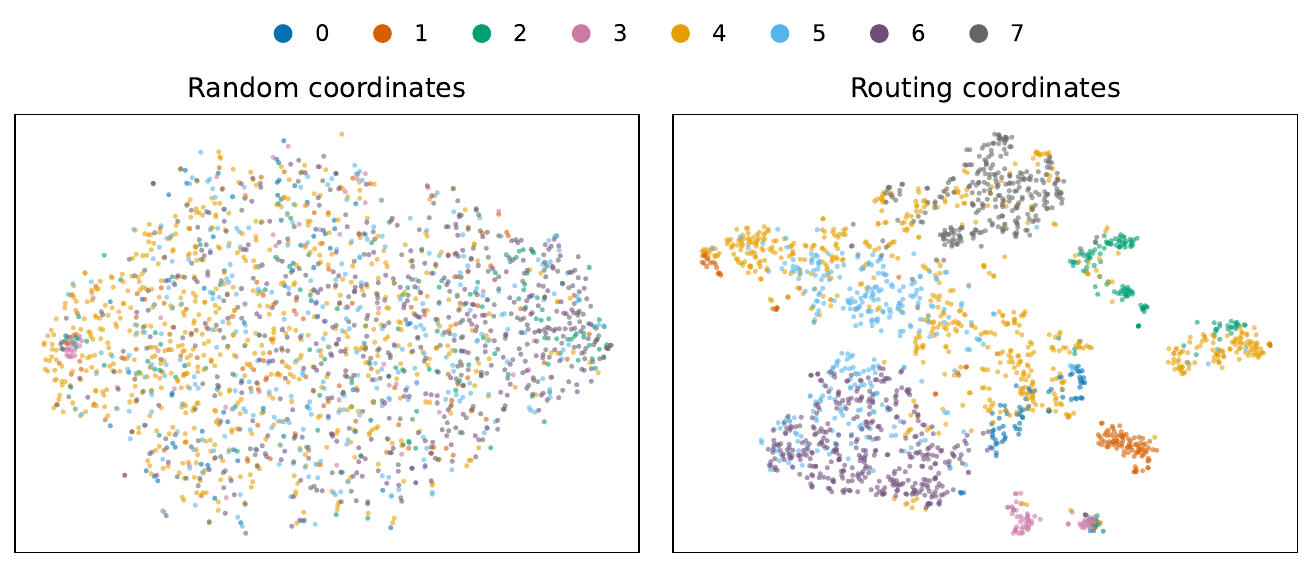}\\[-0.25em]
\scriptsize Layer 7
\end{minipage}

\vspace{0.12em}

\begin{minipage}{0.395\textwidth}
\centering
\includegraphics[width=\linewidth]{Figures/appendix/routing_geometry_post_tsne/post_training_tsne_layer_8.pdf}\\[-0.25em]
\scriptsize Layer 8
\end{minipage}
\begin{minipage}{0.395\textwidth}
\centering
\includegraphics[width=\linewidth]{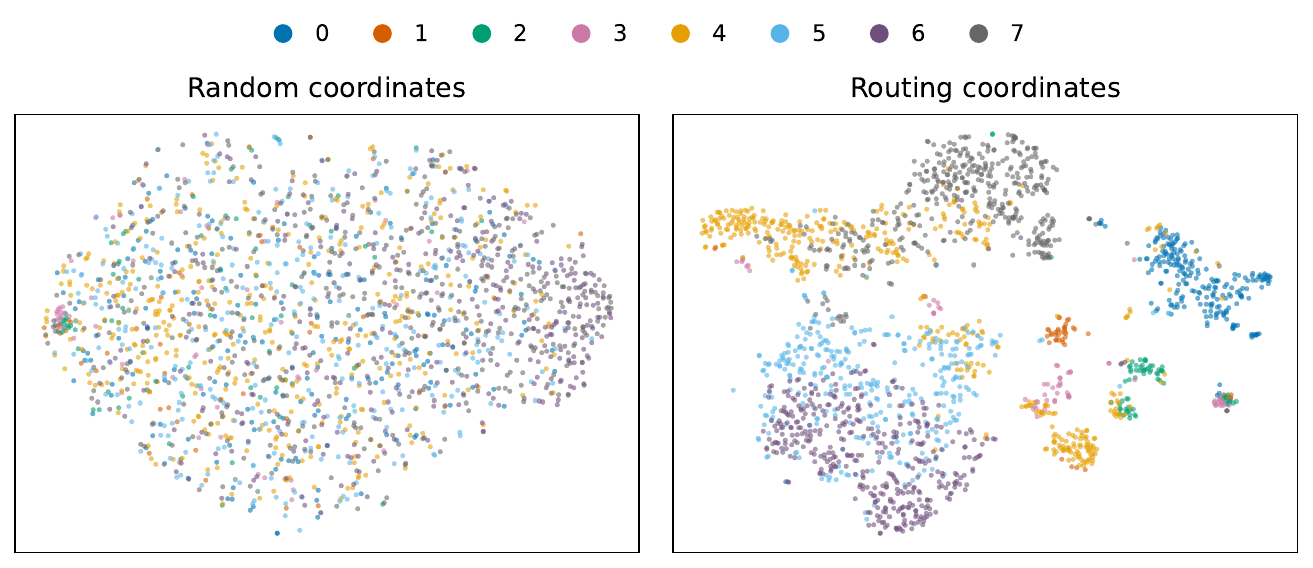}\\[-0.25em]
\scriptsize Layer 9
\end{minipage}

\vspace{0.12em}

\begin{minipage}{0.395\textwidth}
\centering
\includegraphics[width=\linewidth]{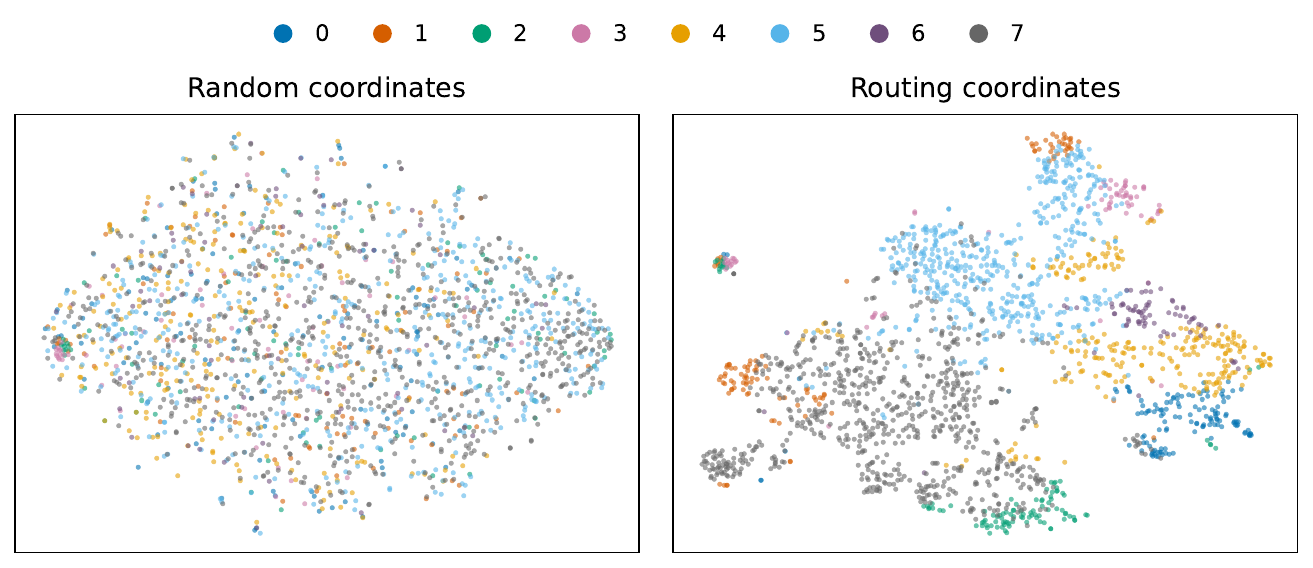}\\[-0.25em]
\scriptsize Layer 10
\end{minipage}
\begin{minipage}{0.395\textwidth}
\centering
\includegraphics[width=\linewidth]{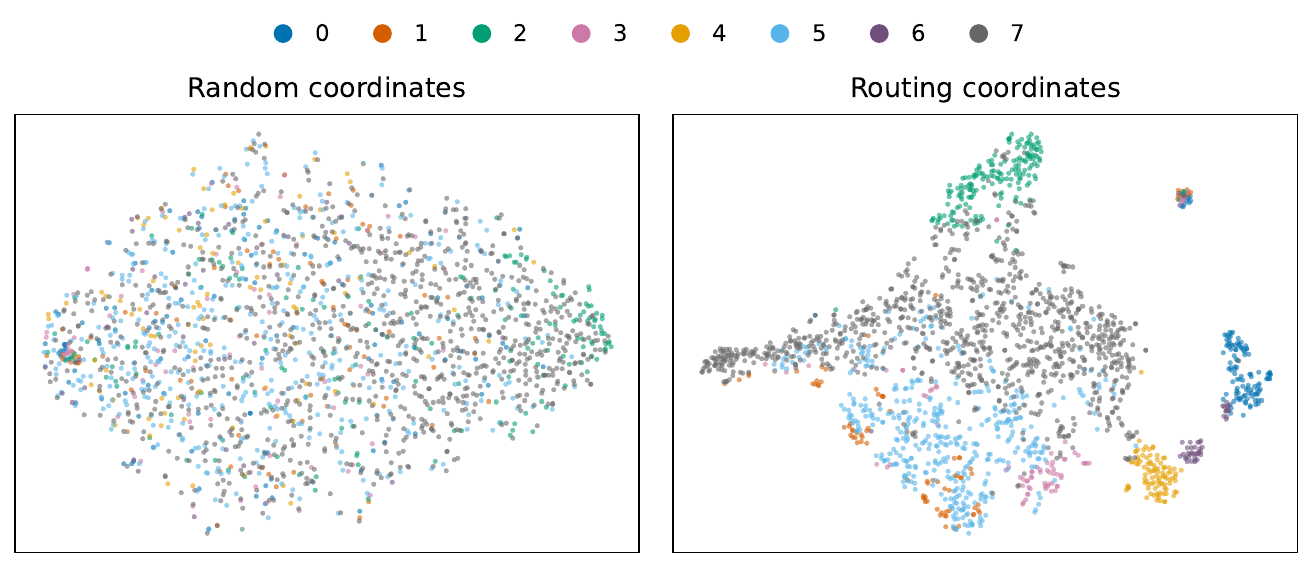}\\[-0.25em]
\scriptsize Layer 11
\end{minipage}
\caption{\textbf{Routing-coordinate geometry after training, last-$N$ coordinates.} Each panel compares random non-routing coordinates with the designated routing coordinates in the trained Self-Routing checkpoint. The earliest layers are less separated, matching the early-layer routing imbalance in Figure~\ref{fig:expert_heatmap}. After these layers, the designated routing coordinates form clearer expert-separated regions than random coordinates, supporting the interpretation that Self-Routing places routing-relevant specialization into the coordinates used for routing.}
\label{fig:app_routing_geometry_lastn}
\end{figure*}

\begin{figure*}[t]
\centering
\begin{minipage}{0.395\textwidth}
\centering
\includegraphics[width=\linewidth]{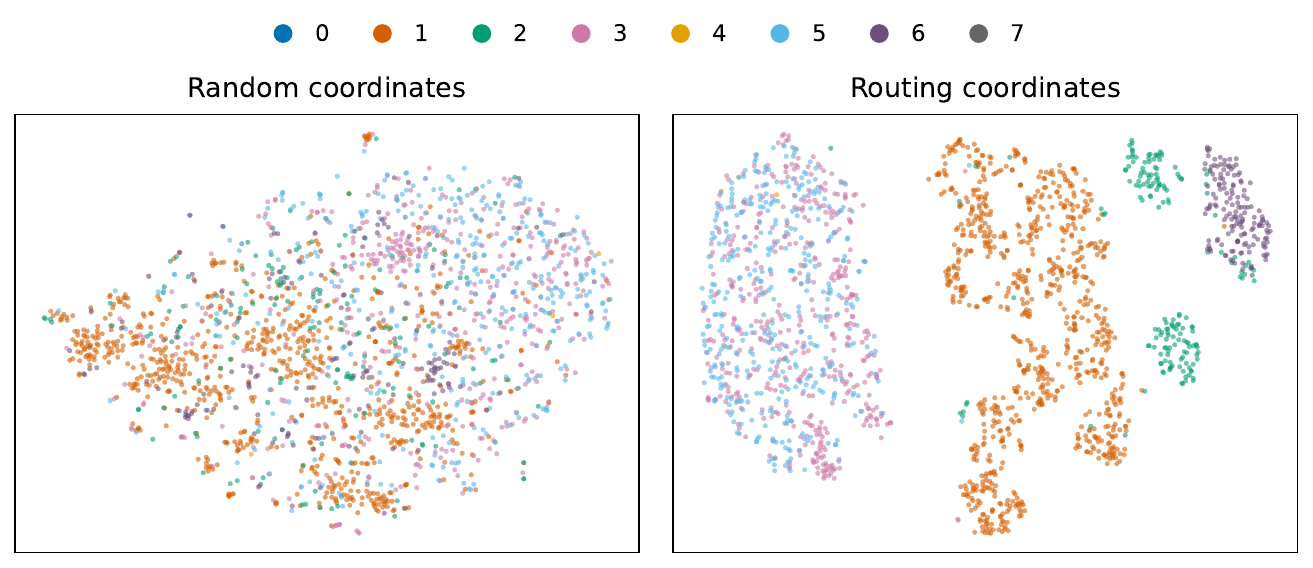}\\[-0.25em]
\scriptsize Layer 0
\end{minipage}
\begin{minipage}{0.395\textwidth}
\centering
\includegraphics[width=\linewidth]{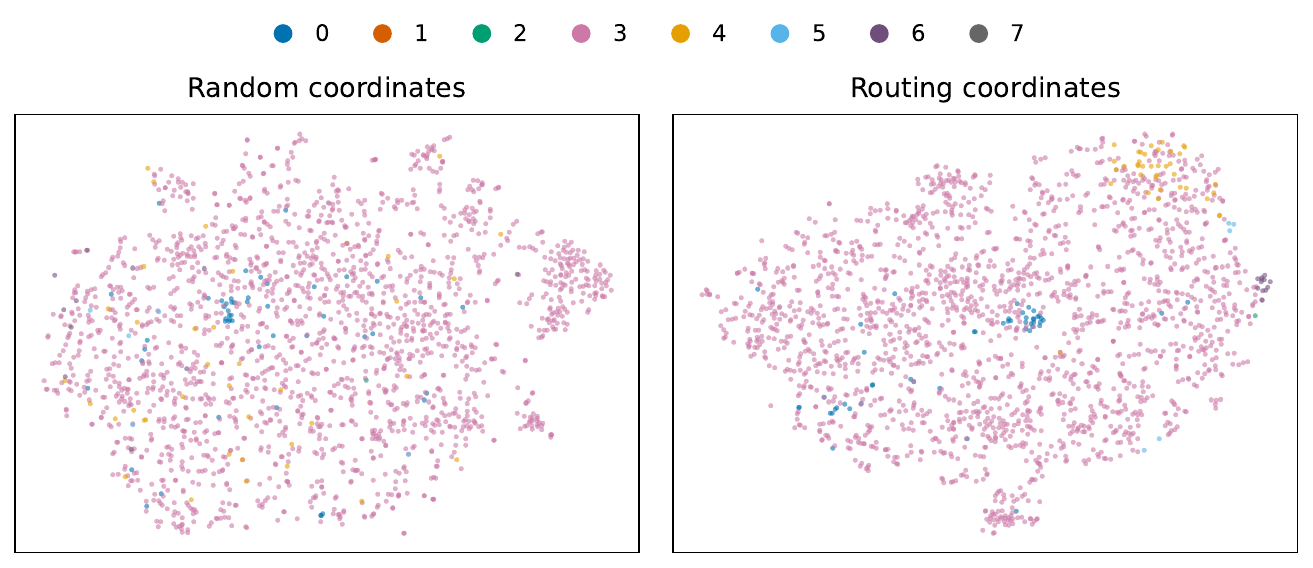}\\[-0.25em]
\scriptsize Layer 1
\end{minipage}

\vspace{0.12em}

\begin{minipage}{0.395\textwidth}
\centering
\includegraphics[width=\linewidth]{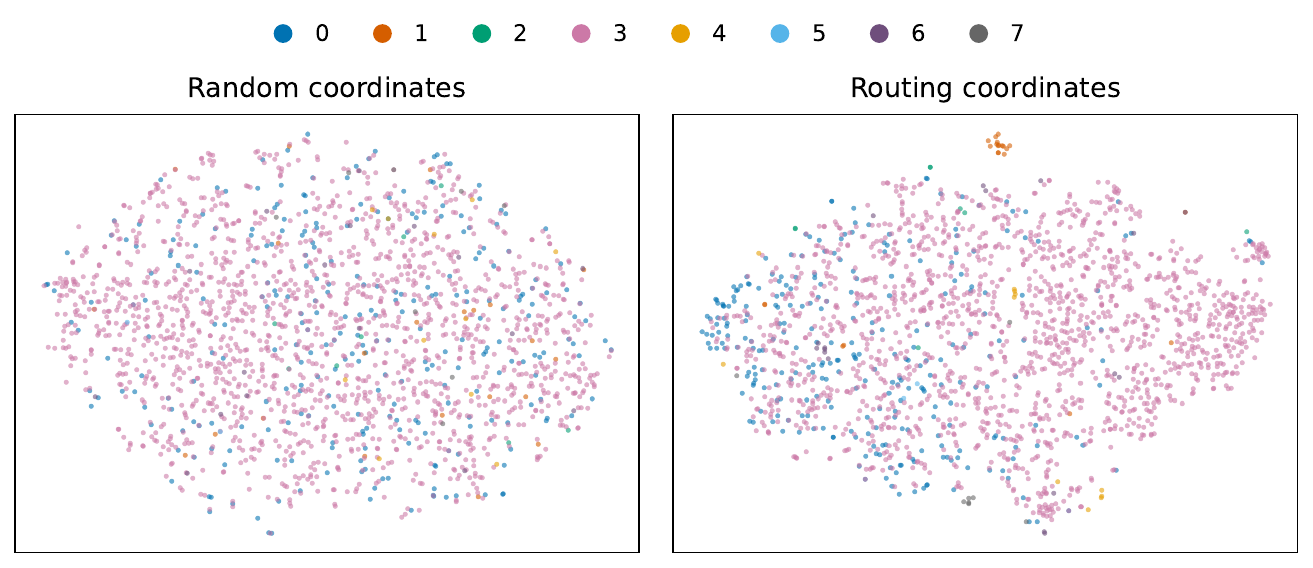}\\[-0.25em]
\scriptsize Layer 2
\end{minipage}
\begin{minipage}{0.395\textwidth}
\centering
\includegraphics[width=\linewidth]{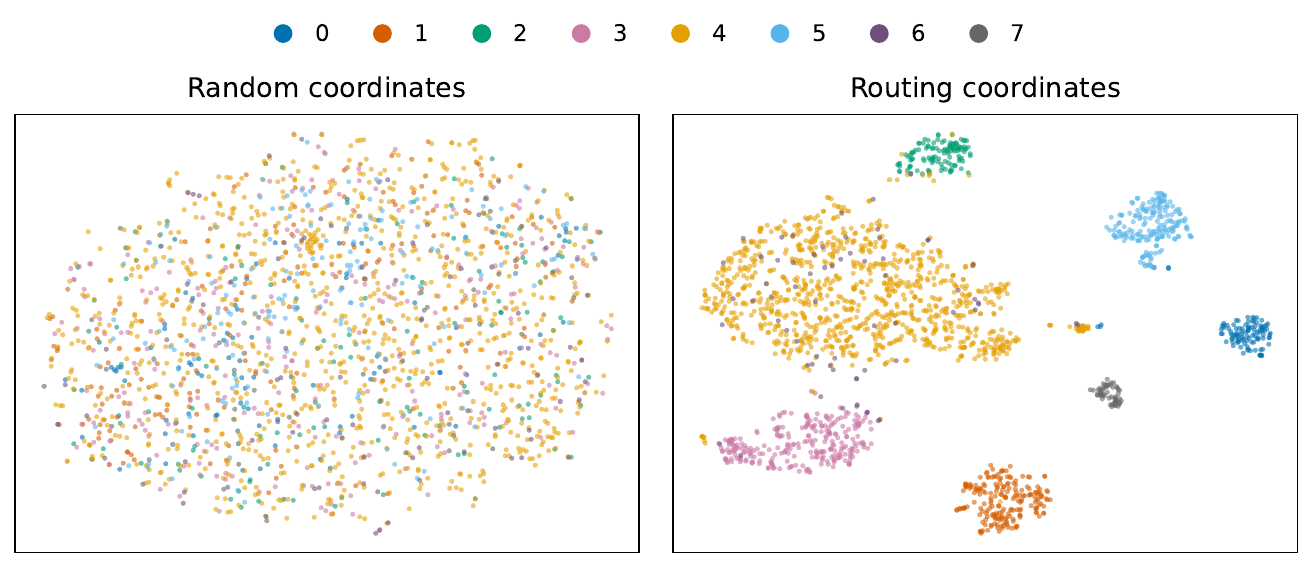}\\[-0.25em]
\scriptsize Layer 3
\end{minipage}

\vspace{0.12em}

\begin{minipage}{0.395\textwidth}
\centering
\includegraphics[width=\linewidth]{Figures/appendix/routing_geometry_self_random_post_tsne/post_training_tsne_layer_4.pdf}\\[-0.25em]
\scriptsize Layer 4
\end{minipage}
\begin{minipage}{0.395\textwidth}
\centering
\includegraphics[width=\linewidth]{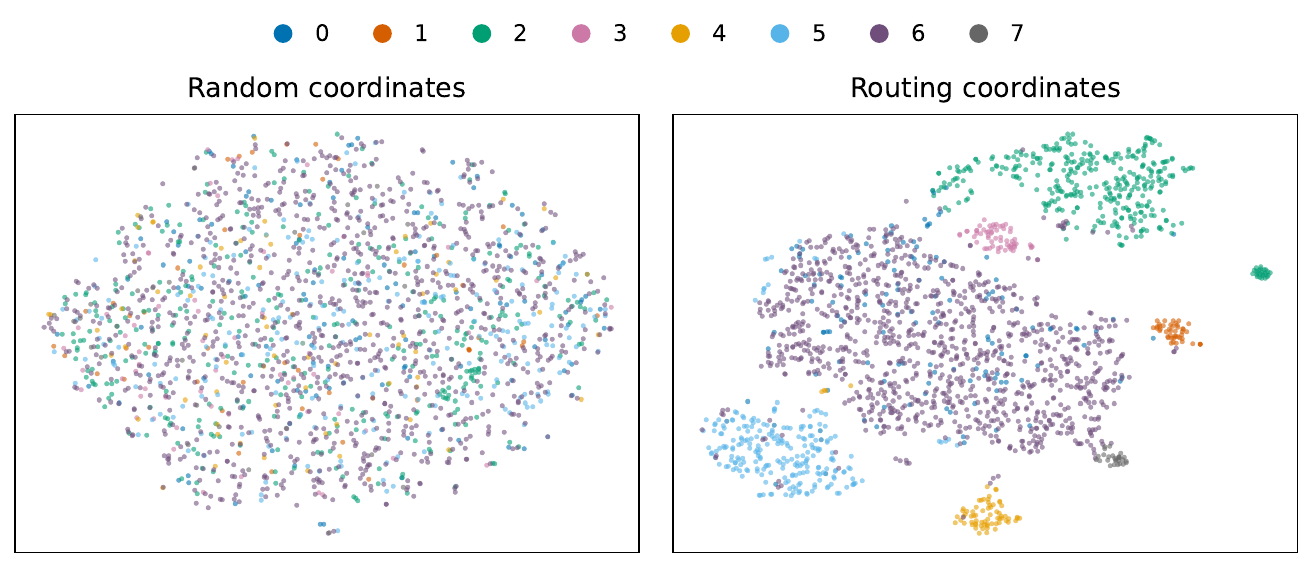}\\[-0.25em]
\scriptsize Layer 5
\end{minipage}

\vspace{0.12em}

\begin{minipage}{0.395\textwidth}
\centering
\includegraphics[width=\linewidth]{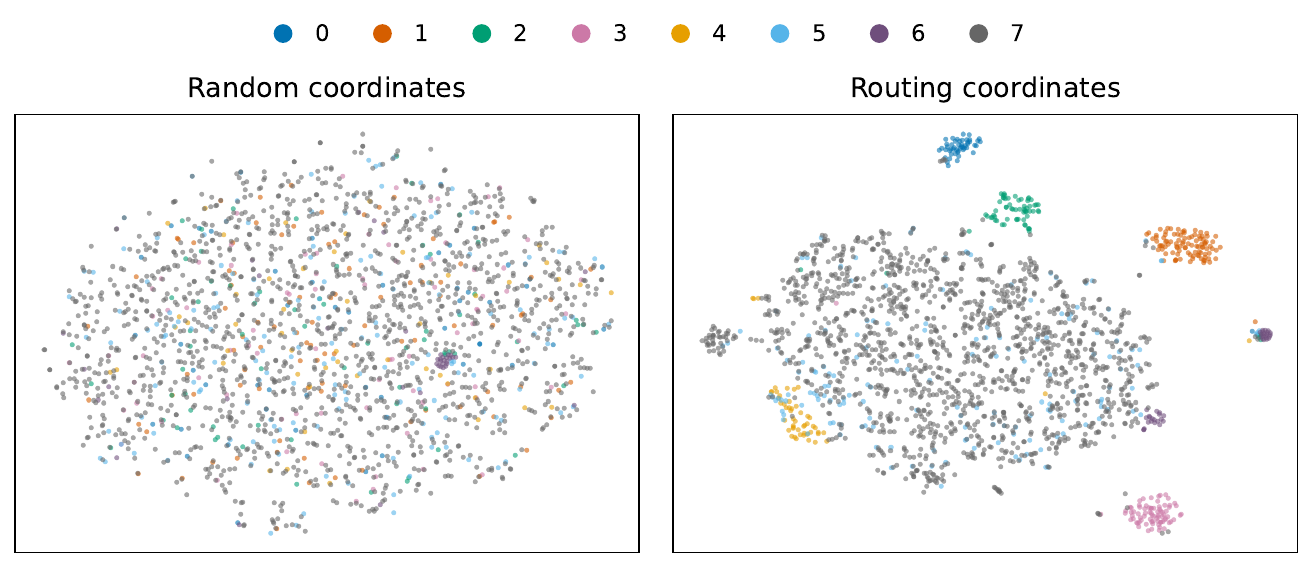}\\[-0.25em]
\scriptsize Layer 6
\end{minipage}
\begin{minipage}{0.395\textwidth}
\centering
\includegraphics[width=\linewidth]{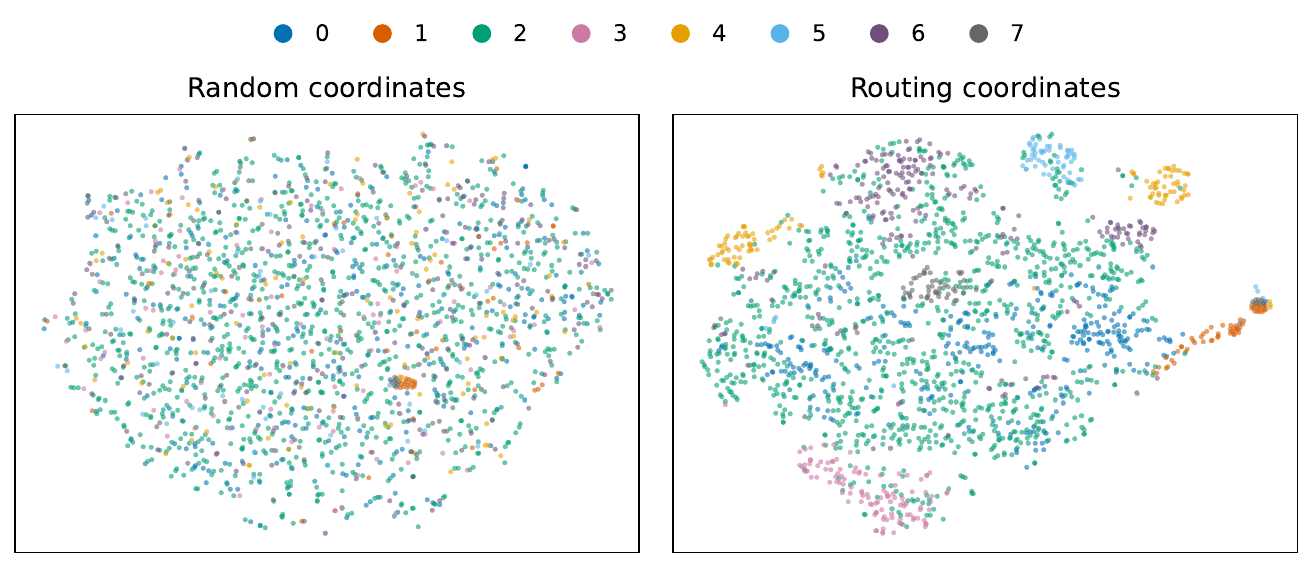}\\[-0.25em]
\scriptsize Layer 7
\end{minipage}

\vspace{0.12em}

\begin{minipage}{0.395\textwidth}
\centering
\includegraphics[width=\linewidth]{Figures/appendix/routing_geometry_self_random_post_tsne/post_training_tsne_layer_8.pdf}\\[-0.25em]
\scriptsize Layer 8
\end{minipage}
\begin{minipage}{0.395\textwidth}
\centering
\includegraphics[width=\linewidth]{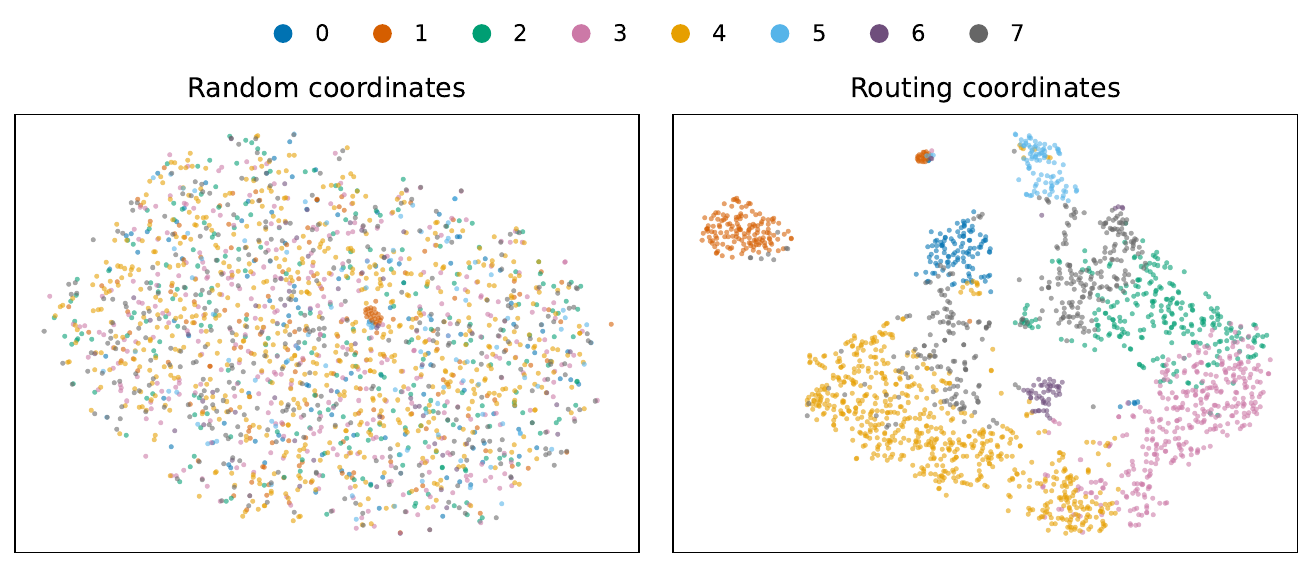}\\[-0.25em]
\scriptsize Layer 9
\end{minipage}

\vspace{0.12em}

\begin{minipage}{0.395\textwidth}
\centering
\includegraphics[width=\linewidth]{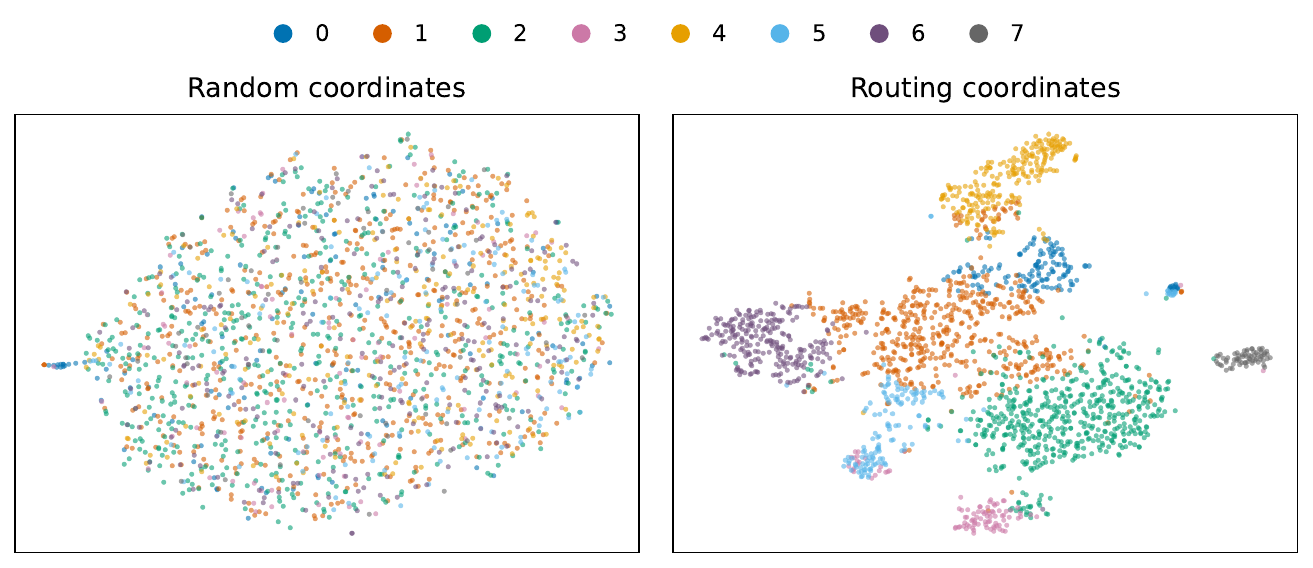}\\[-0.25em]
\scriptsize Layer 10
\end{minipage}
\begin{minipage}{0.395\textwidth}
\centering
\includegraphics[width=\linewidth]{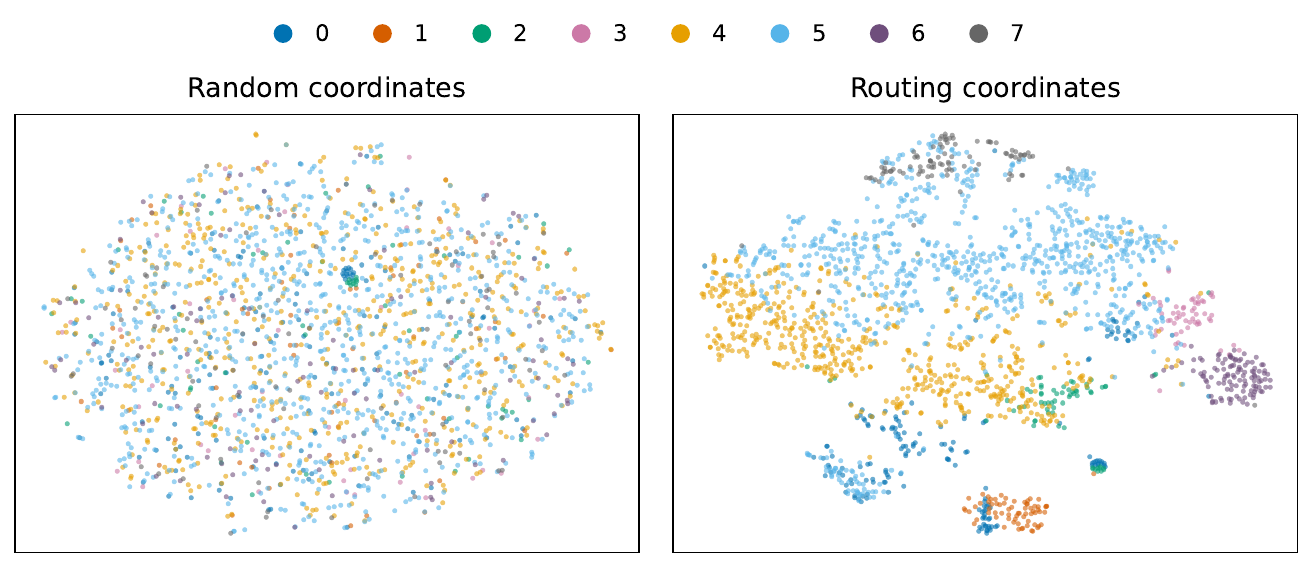}\\[-0.25em]
\scriptsize Layer 11
\end{minipage}
\caption{\textbf{Routing-coordinate geometry after training, random coordinates.} The routing coordinates were sampled randomly at initialization and then kept fixed. Fixed random routing coordinates can also support organized expert-separated regions, showing that routing-relevant specialization can be placed into a designated coordinate subset even when that subset is not the last $N$ hidden dimensions. This reinforces the coordinate-choice ablation in Table~\ref{tab:coordinate_choice}.}
\label{fig:app_routing_geometry_random}
\end{figure*}

Figures~\ref{fig:app_routing_geometry_lastn} and~\ref{fig:app_routing_geometry_random} visualize the LN-normalized routing coordinates with t-SNE. Each panel compares a randomly chosen non-routing coordinate subset with the routing coordinates in the same trained checkpoint. Points are colored by the top-1 routed expert. The routing-coordinate panels form clearer expert-separated regions than the random non-routing coordinates. This indicates that the model learns to place routing-relevant specialization into the designated coordinates, rather than the routing structure simply arising from generic LayerNorm-normalized activations.

\subsection{Routing Analysis}
\label{app:routing_analysis}

\begin{figure*}[p]
\centering
\includegraphics[width=0.94\textwidth,height=0.38\textheight,keepaspectratio]{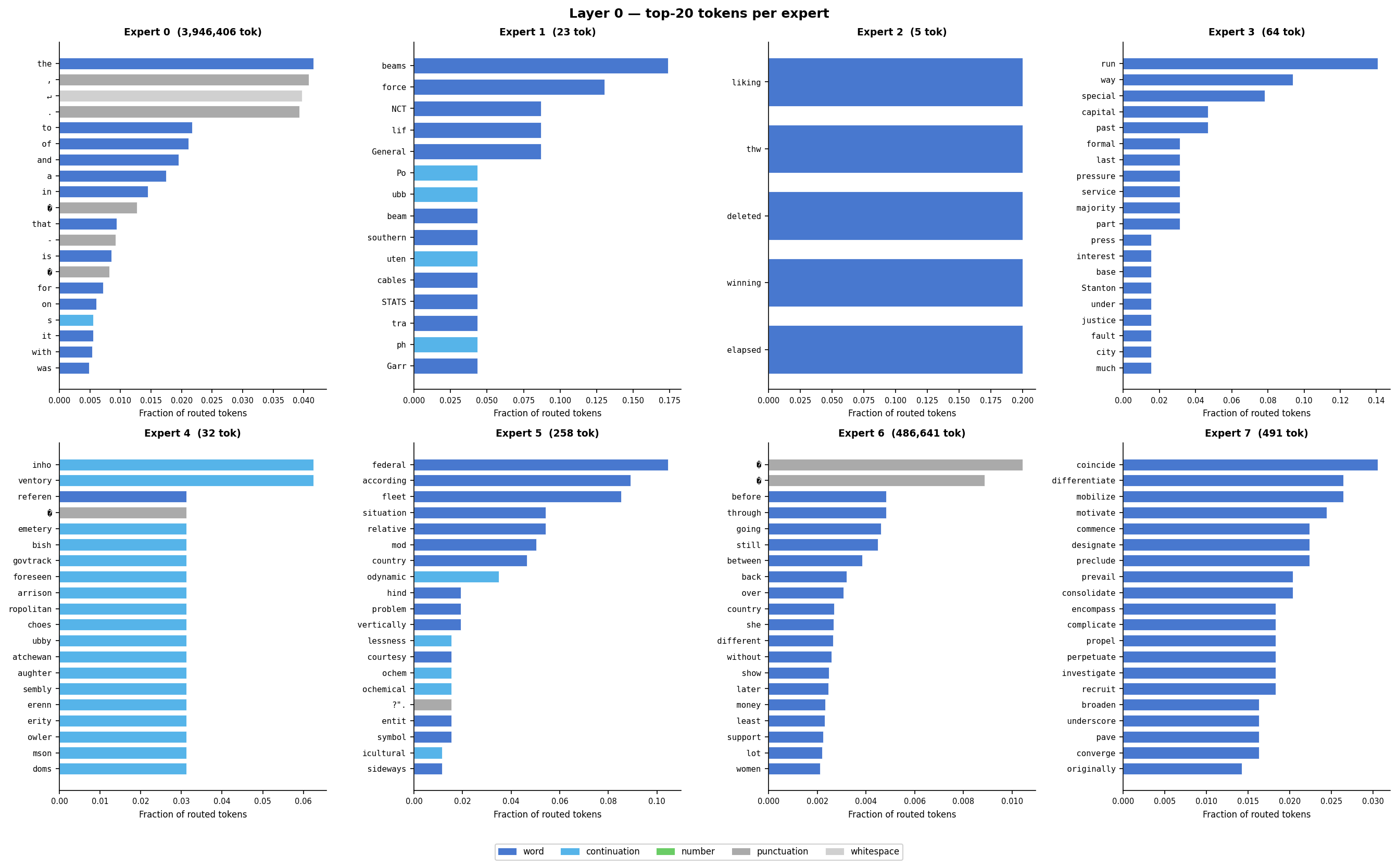}

\vspace{0.2em}

\includegraphics[width=0.94\textwidth,height=0.38\textheight,keepaspectratio]{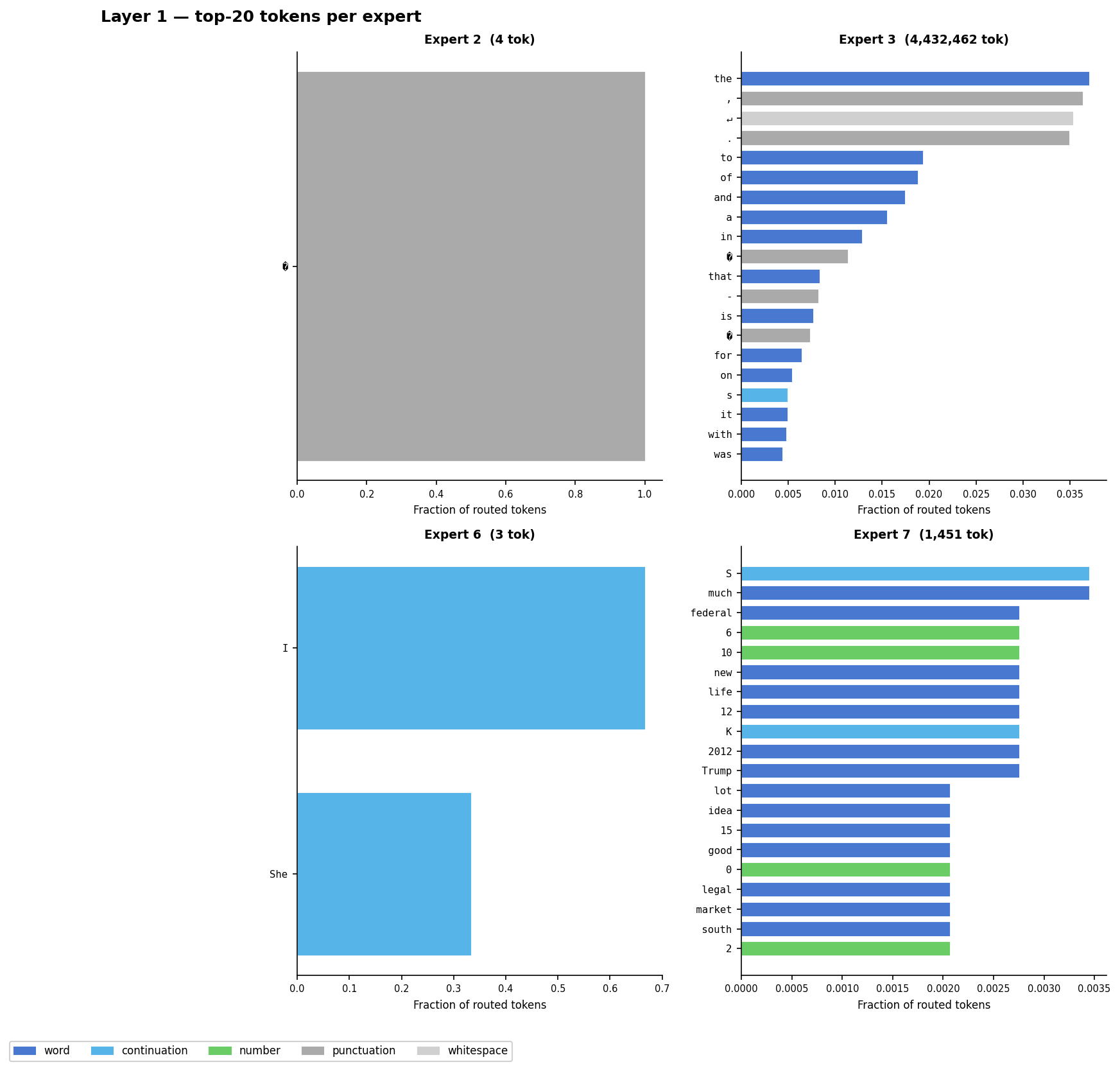}
\caption{\textbf{Top-20 routed tokens per expert, layers 0--1.}}
\label{fig:app_token_affinity_layers_00_01}
\end{figure*}

\begin{figure*}[p]
\centering
\includegraphics[width=0.94\textwidth,height=0.38\textheight,keepaspectratio]{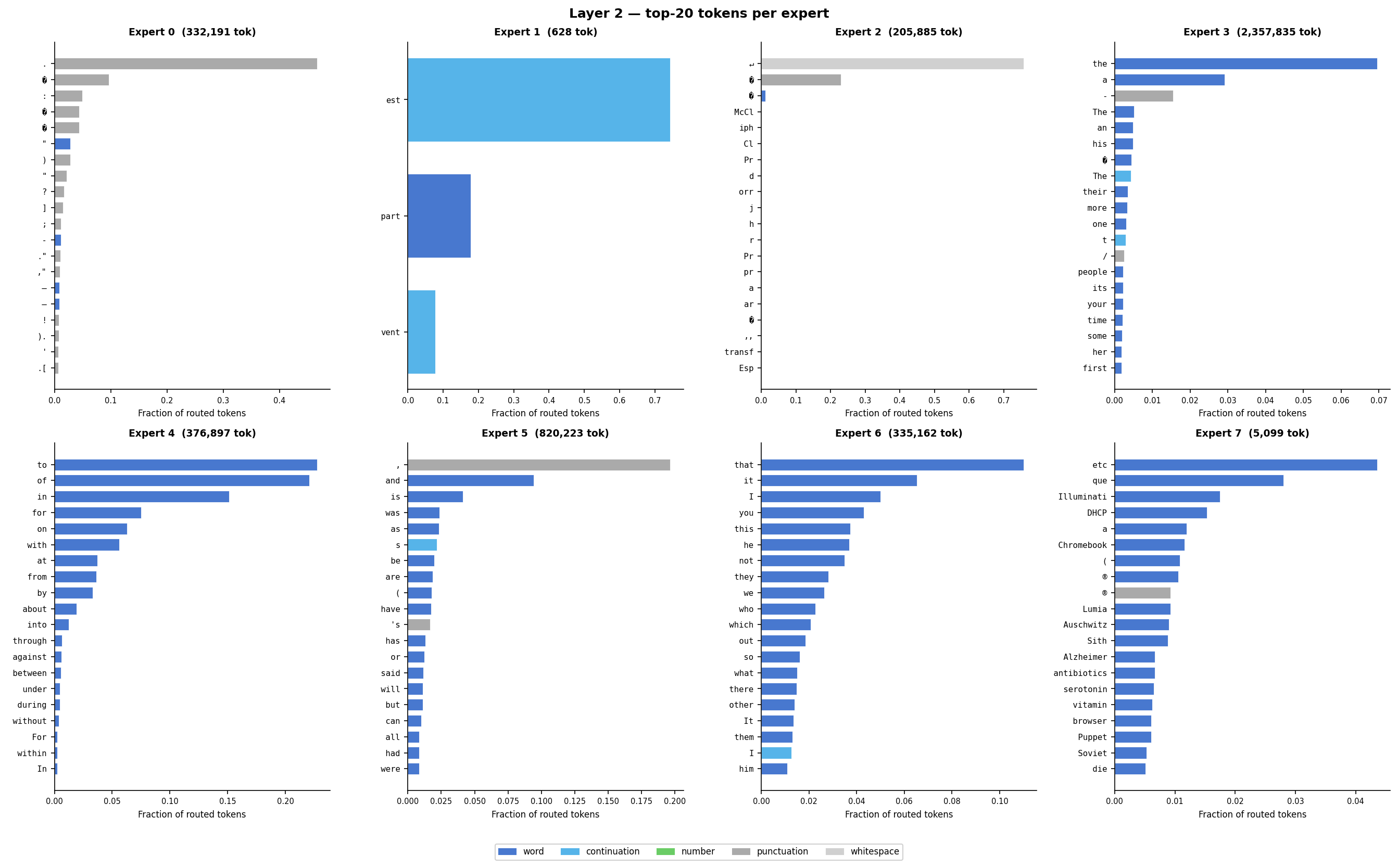}

\vspace{0.2em}

\includegraphics[width=0.94\textwidth,height=0.38\textheight,keepaspectratio]{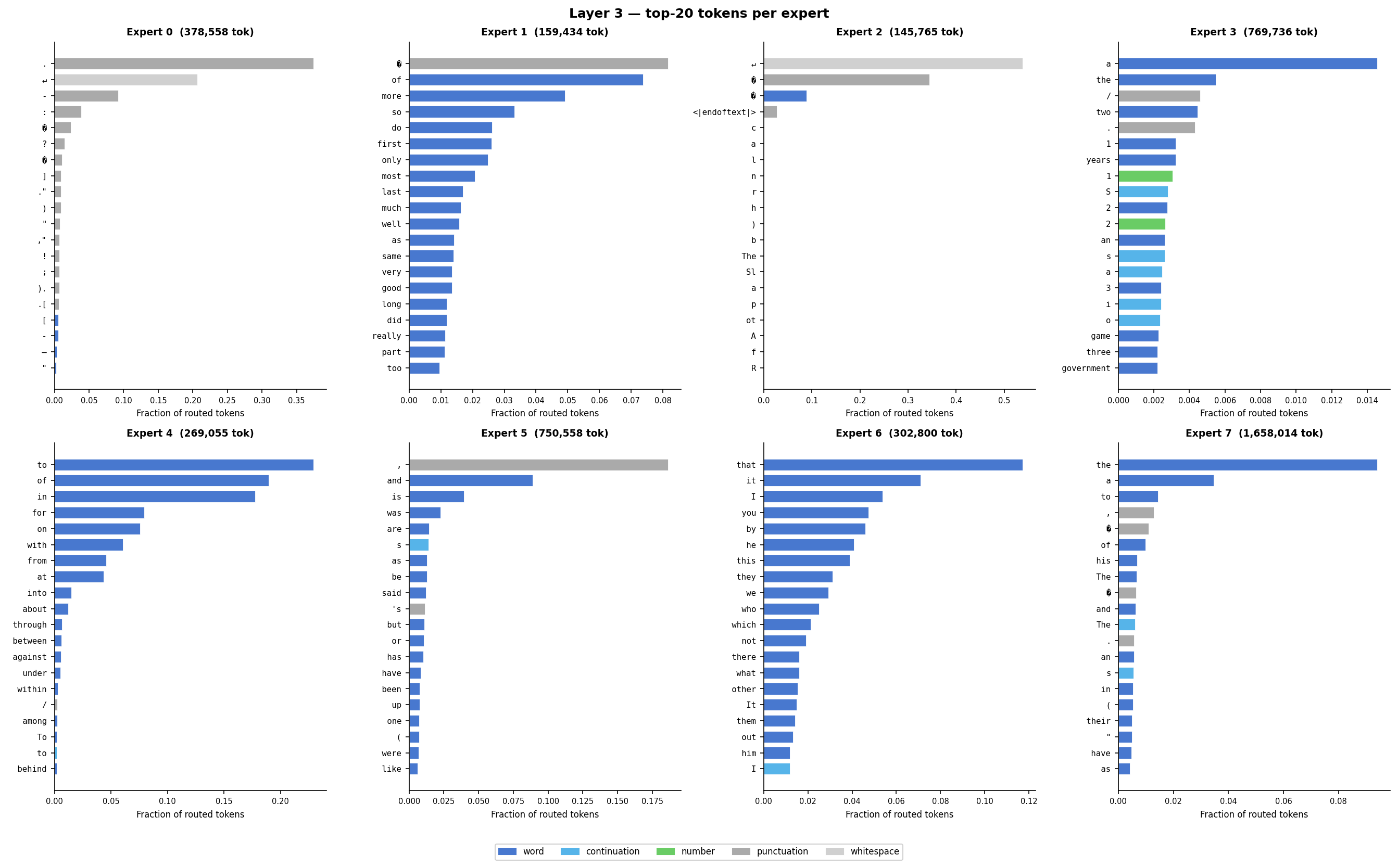}
\caption{\textbf{Top-20 routed tokens per expert, layers 2--3.}}
\label{fig:app_token_affinity_layers_02_03}
\end{figure*}

\begin{figure*}[p]
\centering
\includegraphics[width=0.94\textwidth,height=0.38\textheight,keepaspectratio]{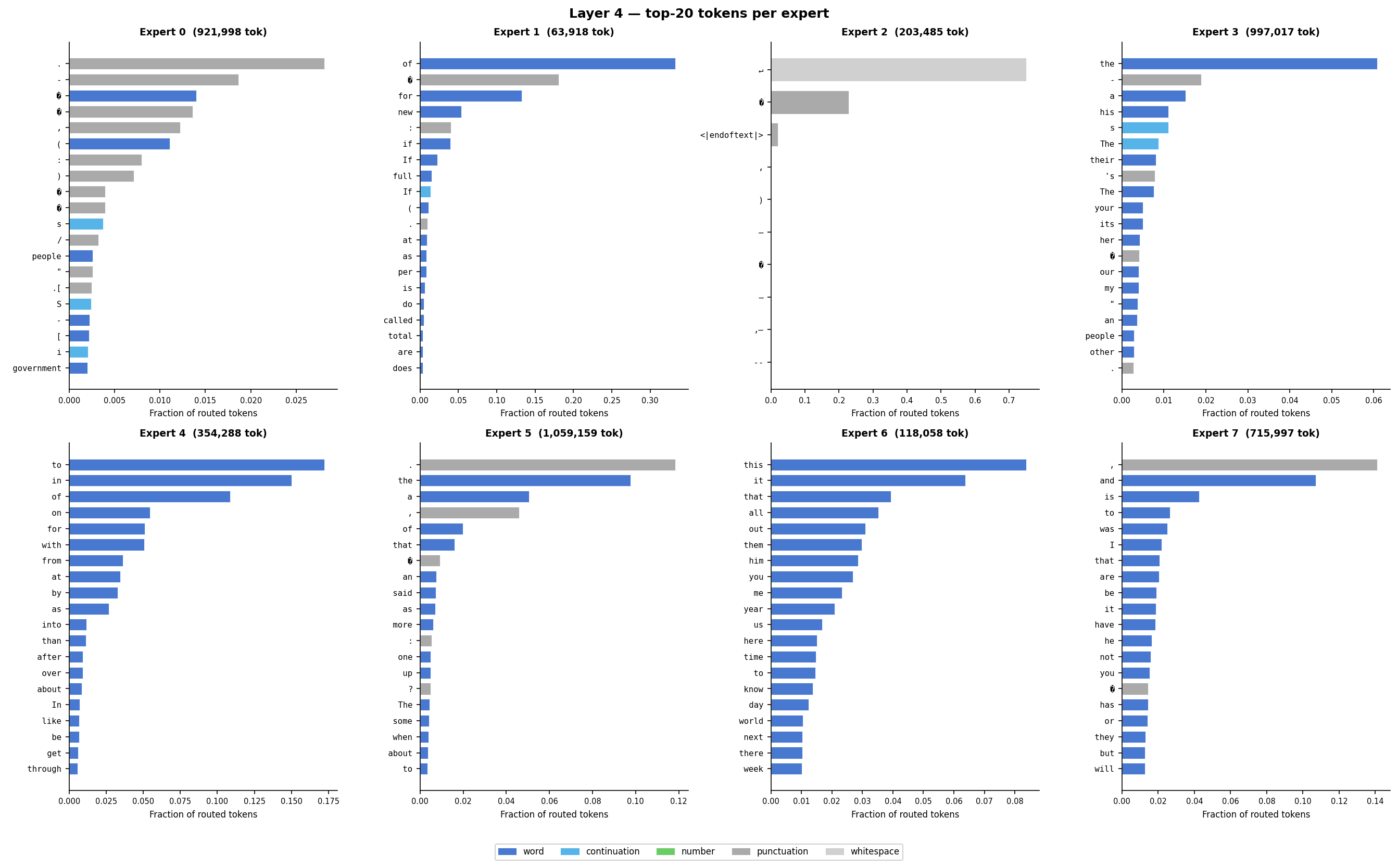}

\vspace{0.2em}

\includegraphics[width=0.94\textwidth,height=0.38\textheight,keepaspectratio]{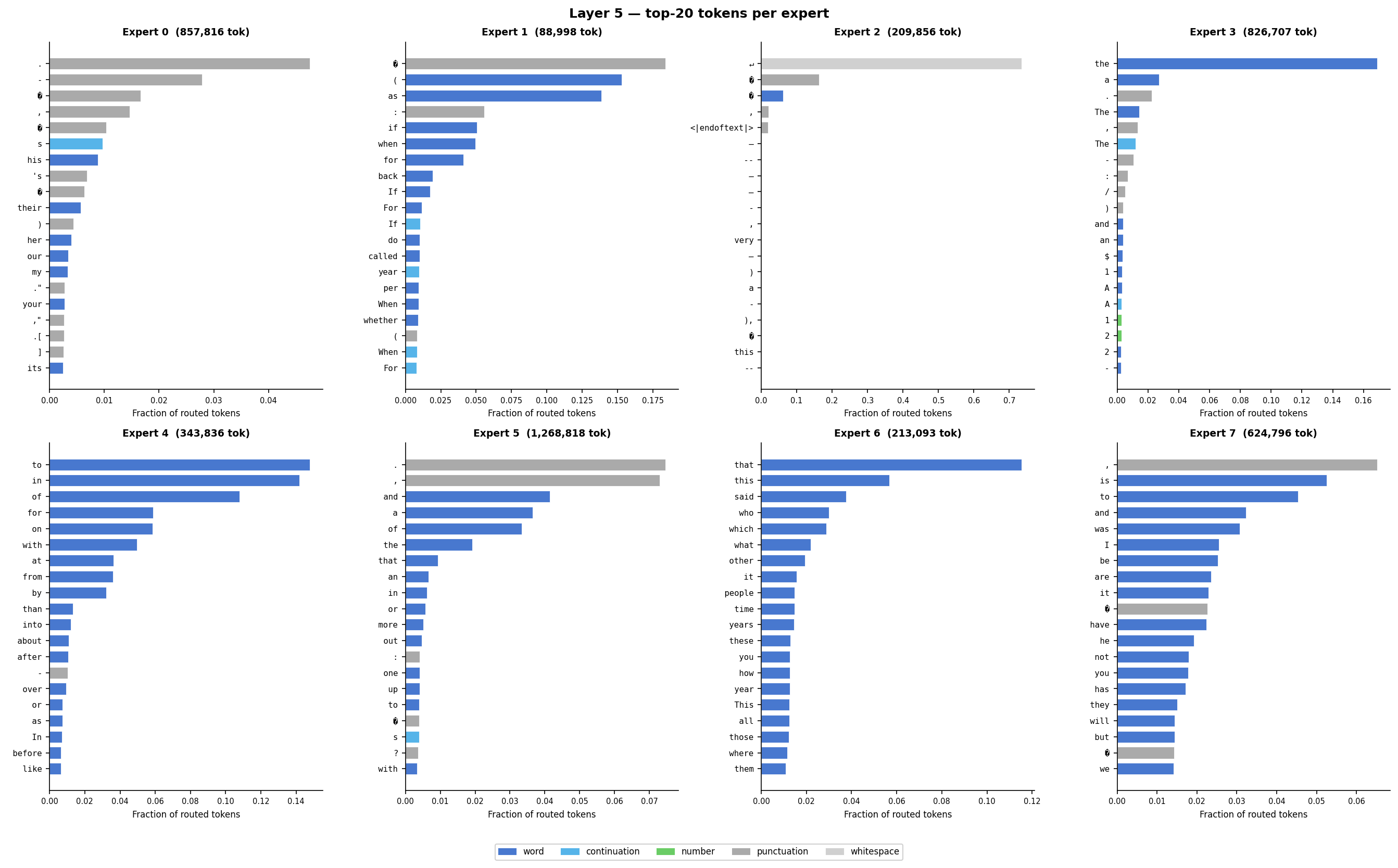}
\caption{\textbf{Top-20 routed tokens per expert, layers 4--5.}}
\label{fig:app_token_affinity_layers_04_05}
\end{figure*}

\begin{figure*}[p]
\centering
\includegraphics[width=0.94\textwidth,height=0.38\textheight,keepaspectratio]{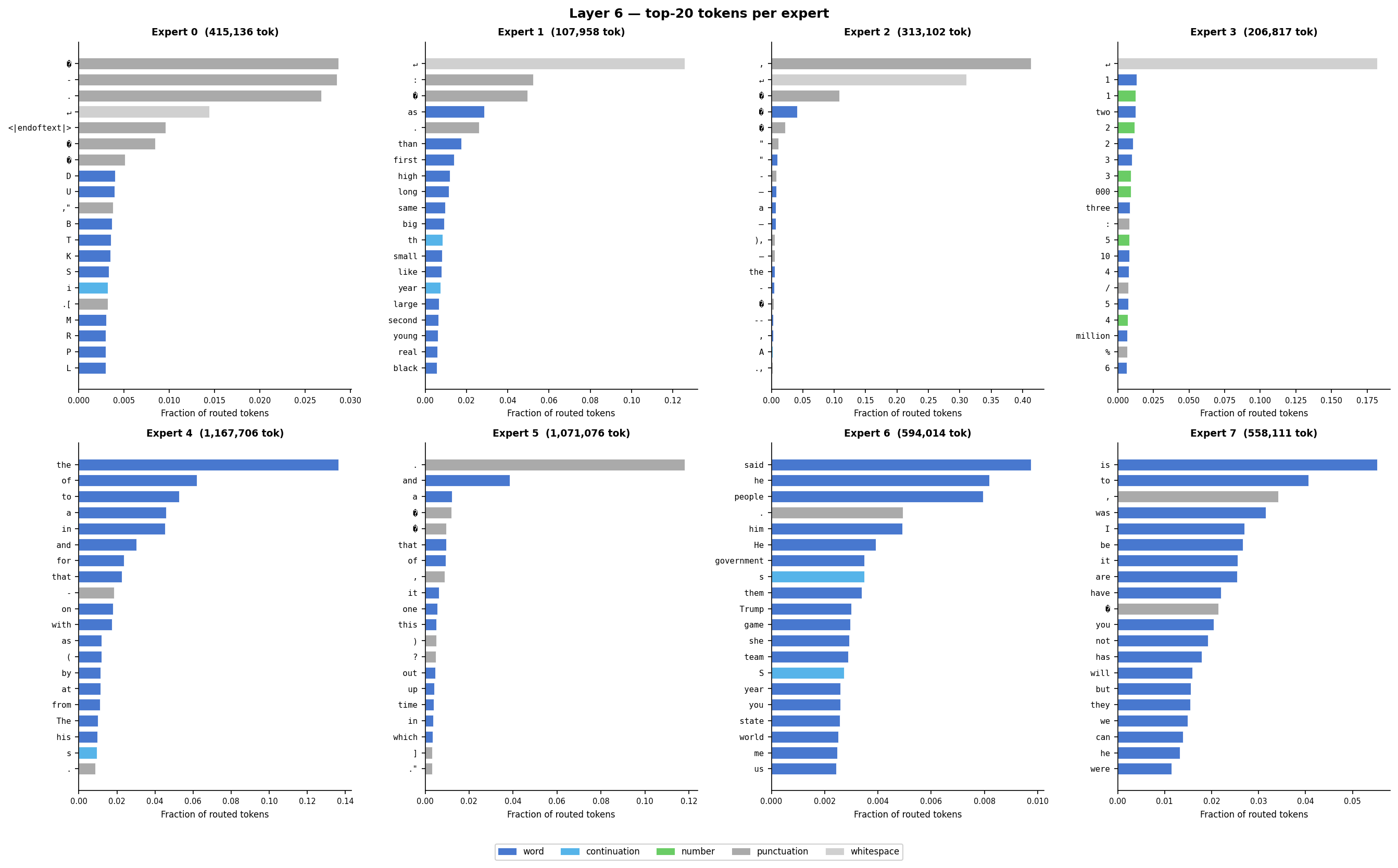}

\vspace{0.2em}

\includegraphics[width=0.94\textwidth,height=0.38\textheight,keepaspectratio]{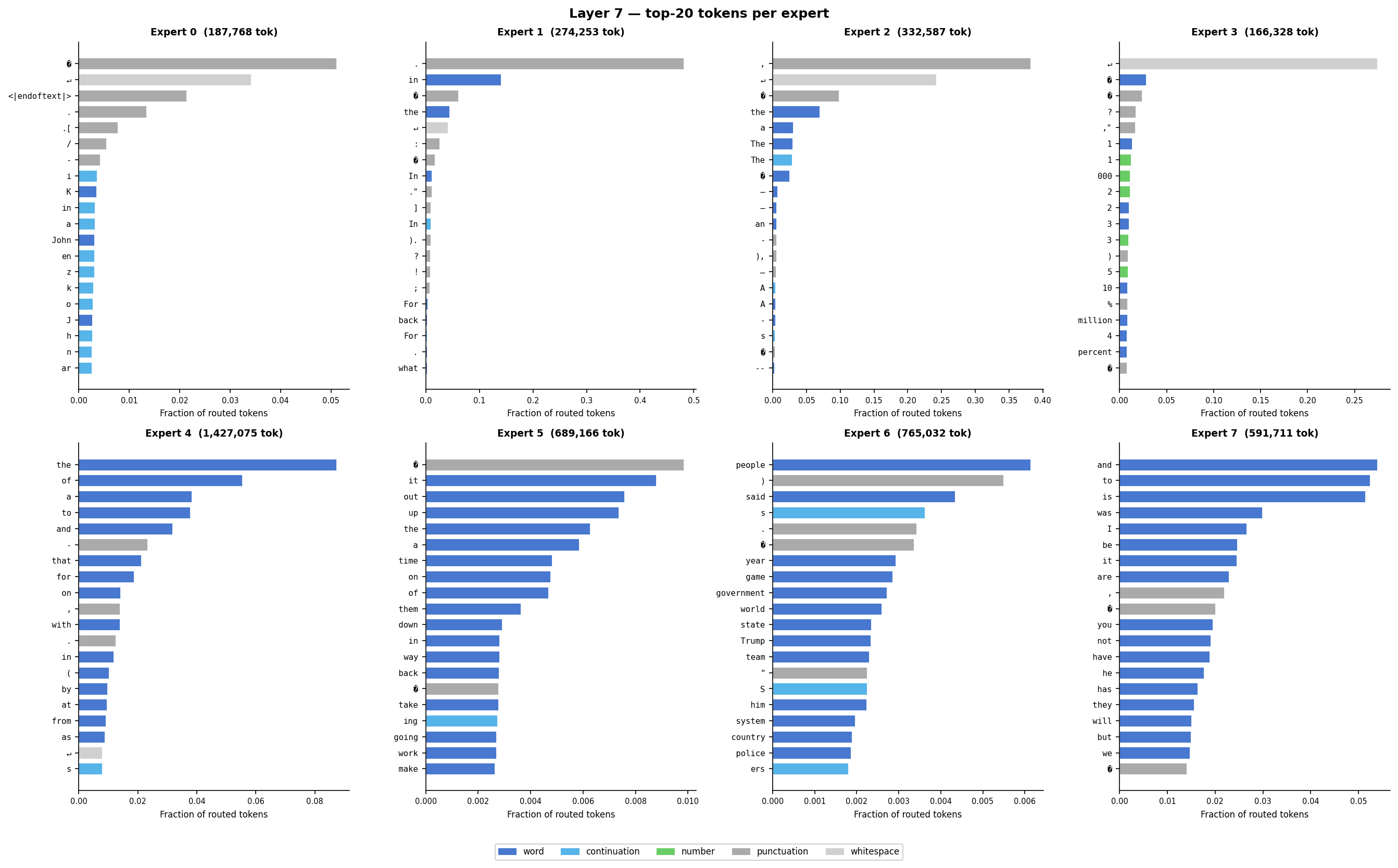}
\caption{\textbf{Top-20 routed tokens per expert, layers 6--7.}}
\label{fig:app_token_affinity_layers_06_07}
\end{figure*}

\begin{figure*}[p]
\centering
\includegraphics[width=0.94\textwidth,height=0.38\textheight,keepaspectratio]{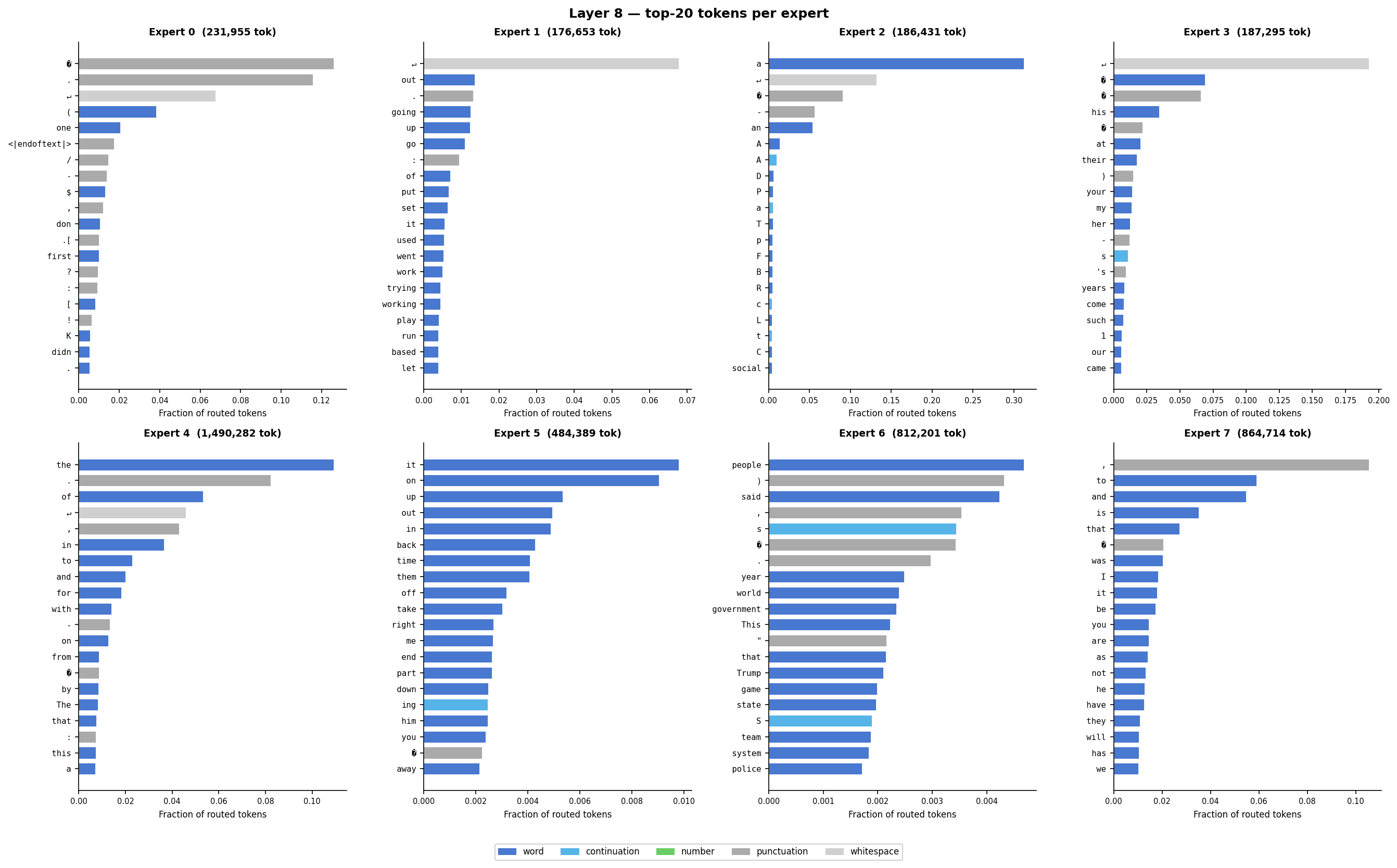}

\vspace{0.2em}

\includegraphics[width=0.94\textwidth,height=0.38\textheight,keepaspectratio]{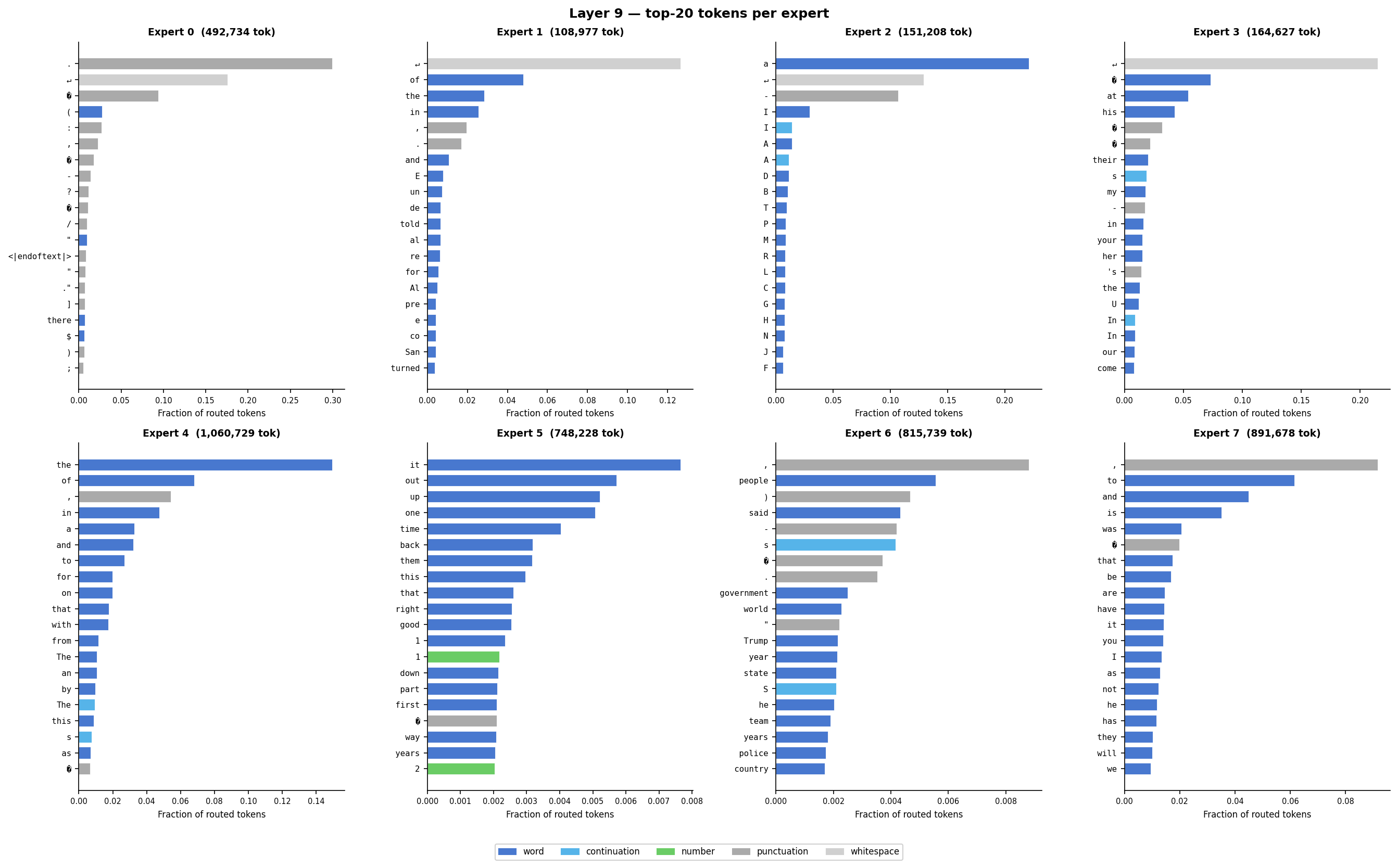}
\caption{\textbf{Top-20 routed tokens per expert, layers 8--9.}}
\label{fig:app_token_affinity_layers_08_09}
\end{figure*}

\begin{figure*}[p]
\centering
\includegraphics[width=0.94\textwidth,height=0.38\textheight,keepaspectratio]{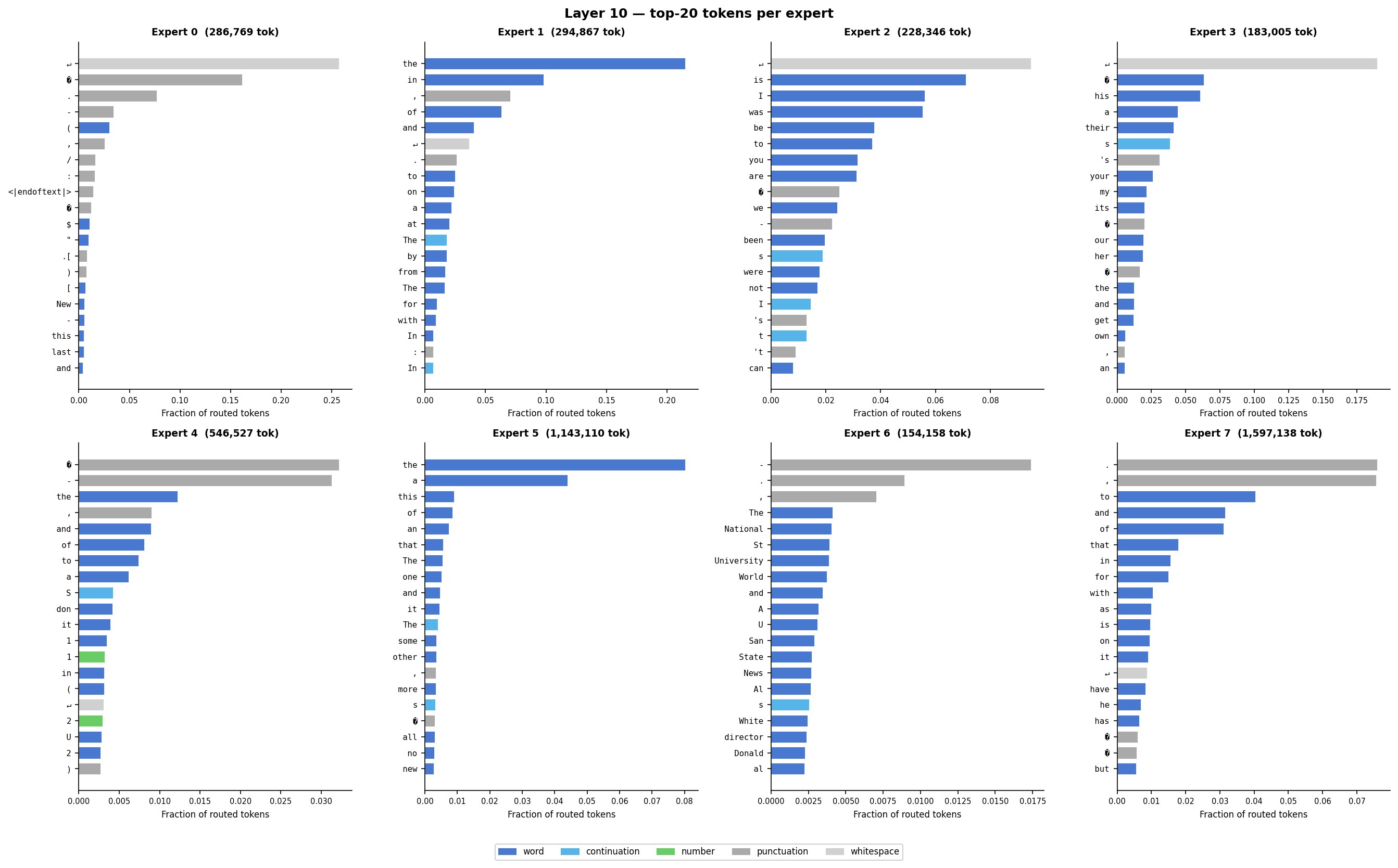}

\vspace{0.2em}

\includegraphics[width=0.94\textwidth,height=0.38\textheight,keepaspectratio]{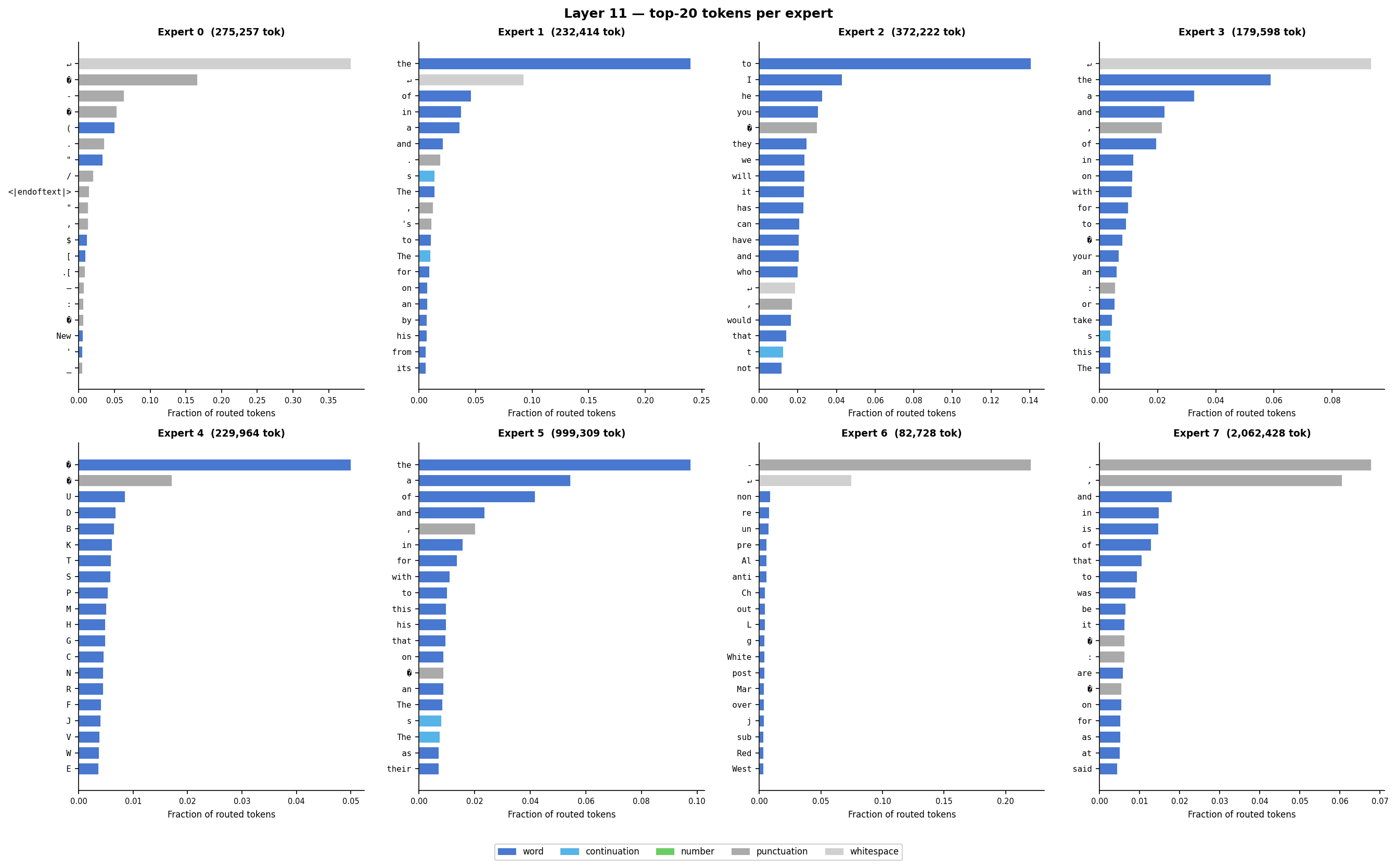}
\caption{\textbf{Top-20 routed tokens per expert, layers 10--11.}}
\label{fig:app_token_affinity_layers_10_11}
\end{figure*}

We analyze the OpenWebText validation set, which contains approximately 4.5M token positions. For each MoE layer, every token position is assigned to its top-1 expert, and for each expert we plot the 20 token IDs routed to that expert most often. Thus, each top-20 list is selected from all routed assignments for that expert, often hundreds of thousands or millions of token assignments, rather than from a small set of examples. The results show expert-specific token patterns. Early layers are noisier; from the middle layers onward, stable patterns emerge around punctuation and document-boundary tokens, relation and function words, numeric tokens, local continuation tokens, and late-layer prefix-like subwords. Results are shown in Figures~\ref{fig:app_token_affinity_layers_00_01}--\ref{fig:app_token_affinity_layers_10_11}.

\subsection{Training Curves}
\label{app:training_curves}

Across the language-modeling and ImageNet runs, the Self-Routing and learned
router curves largely overlap throughout training. The overlap indicates that
Self-Routing follows very similar optimization dynamics to the learned router,
without an obvious stability or convergence disadvantage.

\begin{center}
\centering
\includegraphics[width=\linewidth]{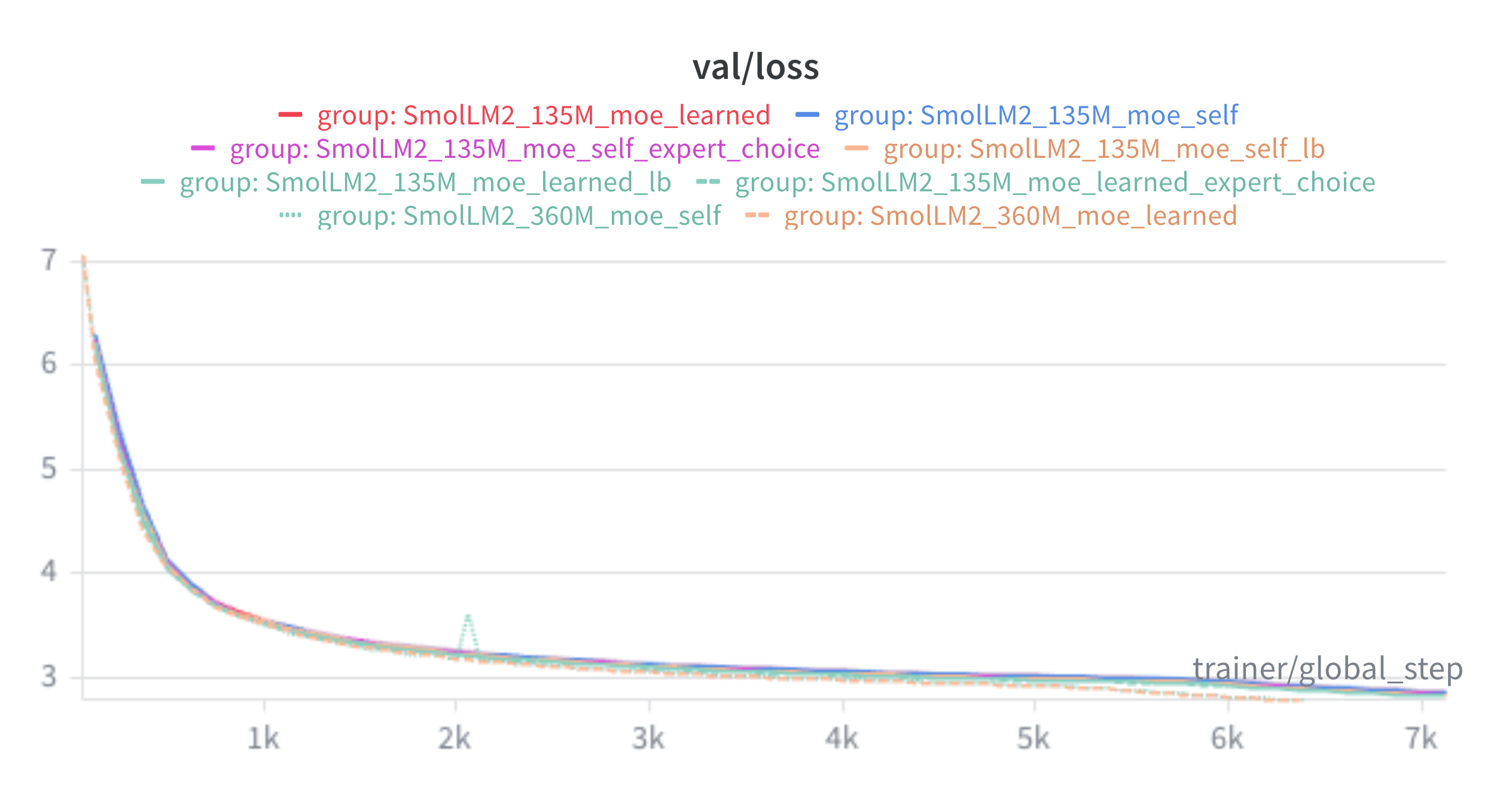}

\vspace{0.2em}

\includegraphics[width=\linewidth]{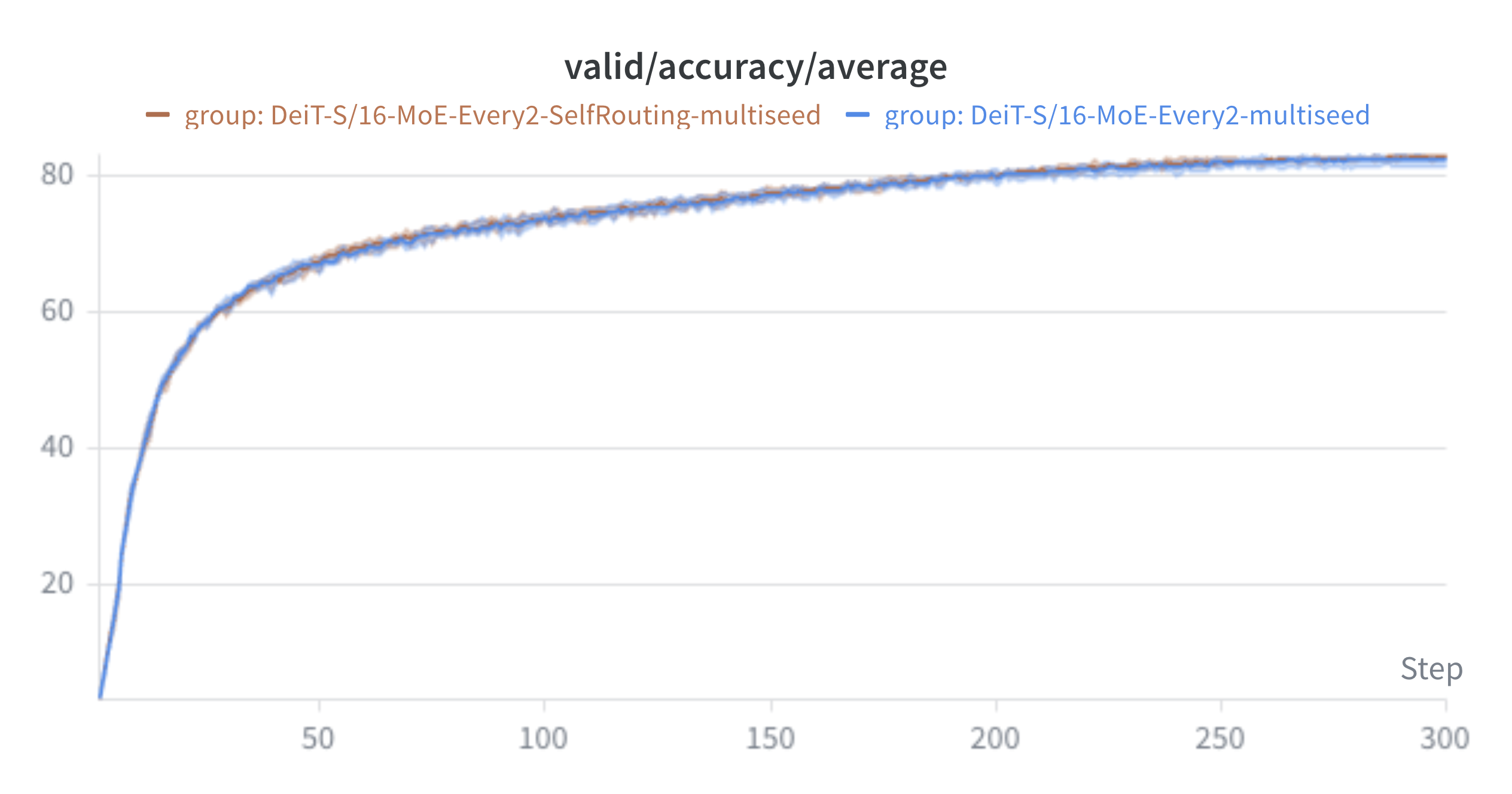}
\captionof{figure}{\textbf{LLaMA validation-loss and ImageNet top-1 validation accuracy curves.}
The corresponding methods follow nearly indistinguishable trajectories.}
\label{fig:app_llama_imagenet_training_curves}
\end{center}

\begin{center}
\centering
\includegraphics[width=\linewidth]{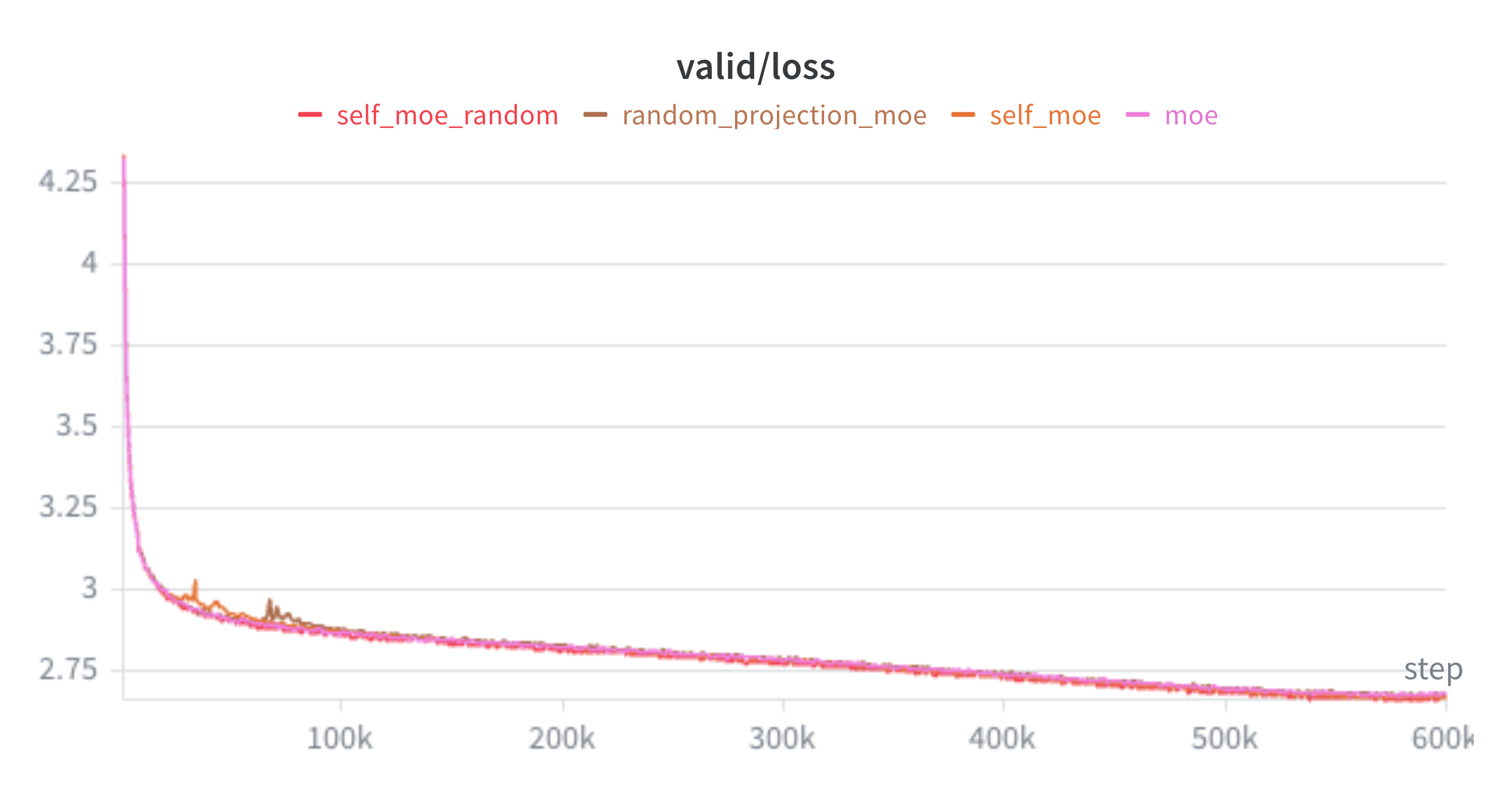}

\vspace{0.2em}

\includegraphics[width=\linewidth]{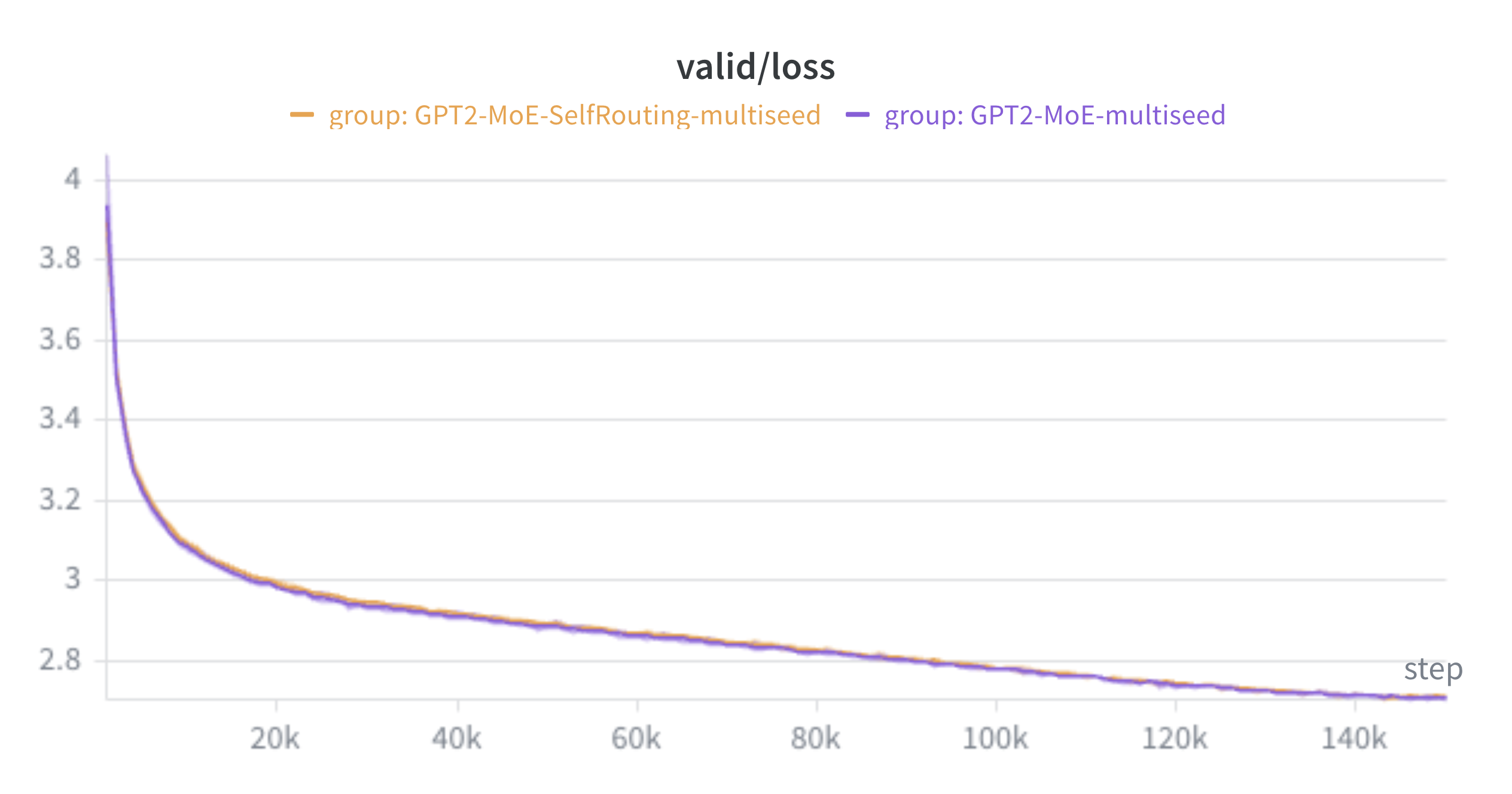}
\captionof{figure}{\textbf{GPT-2 validation-loss curves.}
Self-Routing and learned-router curves largely overlap during training.}
\label{fig:app_gpt2_training_curves}
\end{center}

\end{document}